\def\year{2021}\relax
\documentclass[letterpaper]{article} 
\usepackage{aaai21}  
\usepackage{times}  
\usepackage{helvet} 
\usepackage{courier}  
\usepackage[hyphens]{url}  
\usepackage{graphicx} 
\usepackage{graphicx}  
\usepackage{natbib}  
\usepackage{caption} 
\usepackage{cite} 
\usepackage{float}
\title{Using UNet and PSPNet to explore the reusability principle of CNN parameters }
\author{
Wei Wang
}

\begin{document}

\maketitle

\begin{abstract}
How to reduce the requirement on training dataset size is a hot topic in deep learning community.  One straightforward way is to reuse some pre-trained parameters. Some previous work like Deep transfer learning reuse the model parameters trained for the first task as the starting point for the second task, and semi-supervised learning is trained upon a combination of labeled and unlabeled data. However, the fundamental reason of the success of these methods is unclear. In this paper, the reusability of parameters in each layer of a deep convolutional neural network is experimentally quantified by using a network to do segmentation and auto-encoder task. This paper proves that network parameters can be reused for two reasons: first, the network features are general; Second, there is little difference between the pre-trained parameters and the ideal network parameters. Through the use of parameter replacement and comparison, we demonstrate that reusability is different in BN(Batch Normalization)\cite{bn} layer and Convolution layer and some observations: (1)Running mean and running variance plays an important role than Weight and Bias in BN layer.(2)The weight and bias can be reused in BN layers.(3) The network is very sensitive to the weight of convolutional layer.(4) The bias in Convolution layers are not sensitive, and it can be reused directly.
\end{abstract}

\section{Introduction}
As is known to all, deep learning needs a lot of data for training, but some data are difficult to obtain. In addition, a lot of labeled data is needed in supervision learning, and labeled data requires a lot of human efforts. Therefore, how to use a small amount of data to train and the results can meet our requirements is one of the frontier directions of deep learning.  

There are two ways to solve this problem:  

1.	Transfer learning: A machine learning approach that uses model parameters developed for a task as the starting point for the second model parameter training. Deep transfer learning refers to the reuse of part of the pre-trained network of the original domain, including its network structure and parameters, as part of the deep neural network used in the target domain.  

2.	Semi-supervised learning: Semi-supervised learning is an algorithm combining supervised learning and unsupervised learning. It uses both data with and without labels for learning. At present, the popular practice in deep learning applications is unsupervised pre-training: train the auto-encoder network with all the data, and then take the parameters of the auto-encoder network as the initial parameters and fine-tune them with the labeled data.  

In \cite{2014arXiv1411.1792Y}, the author proposes to use the generalization of neural network for transfer learning. Two factors affecting the transfer learning performance are given: the middle of fragilely co-adapted layers and the specialization of higher layer features.Transfer learning is a method based on generality of features\cite{Caruana,10.5555/3045796.3045800,Bengio2}. Since features are generality, we can first train a set of parameters based on one data set and then transfer to another data set. Semi-supervised learning is a method of using both labeled data and unlabeled data when training data is insufficient. In \cite{Curriculum,Multi-Task,Semi-supervised}, the semi-supervised methods are used for image segmentation.  

At present, both deep transfer learning and semi-supervised deep learning based on the above theory have the same problem: the blindness of parameters reuse selection. Although \cite{2014arXiv1411.1792Y} try to explore the general network layers, but every time they add a network layer they retrain the model. This paper argues that this method cannot explain the generality of network layers, because the network has the adjustment ability, the cause of the results might be: The more layers that are left for network fine-tuning, the stronger the network fine-tuning capability, rather than the more generic layer at the front of the network. The reason why parameters reuse layers can be reused in \cite{Multi-Task,Semi-supervised} are not given.  

This paper believes that both deep transfer learning and semi-supervised deep learning (which is considered as a special kind of transfer learning and can be called self-transfer learning) are essentially the reuse of several layers of parameters with the same network structure of CNN network. Deep transfer learning reuses other data set pre-trained parameters. Semi-supervised learning is the reuse of pre-trained parameters of unlabeled data, which are equivalent to giving a good starting point for target tasks. However, parameter reuse requires two prerequisites: first, the parameters of the reuse layers are general; second, the parameters difference of the corresponding reuse layers are small enough.  

This paper proposes a method to train multiple CNNs on the same or similar data set(Different CNN tasks in the same data set - semi-supervised deep learning method, different data set tasks can be the same or different - deep transfer learning), then replace the network(Requirement: Two or more network models have the same or part of the same network structure.) parameters, compare the results, and find the reusable parameters of each layer of the network. Take image segmentation reuse image auto-encoder parameters in the same data set as an example. The experiment was carried out for the segmentation task, and the reusability of the parameters was determined by the change of the Dice values of the segmentation result. First, a segmentation task model is trained by using the network, and then an image auto-encoder task model is trained. Parameters of the corresponding layers of the segmentation task are replaced by parameters of the image auto-encoding task layer by layer. Don't retrain the model, test the model directly on the validation set, and compare the results with the original result to determine the reusability of the parameters.  

Make the following definition: in the PyTorch environment, the parameters of BN layer Running mean, Running VAR, weight, and bias are respectively called RM, RV, RW, and RB. The convolution layer parameters weight and bias are respectively called W and B.   

PSPNet \cite{psp} and UNet \cite{unet} were used in this paper. The network structure is as follows:

\begin{figure}[h]
\centering
\includegraphics[scale=0.18]{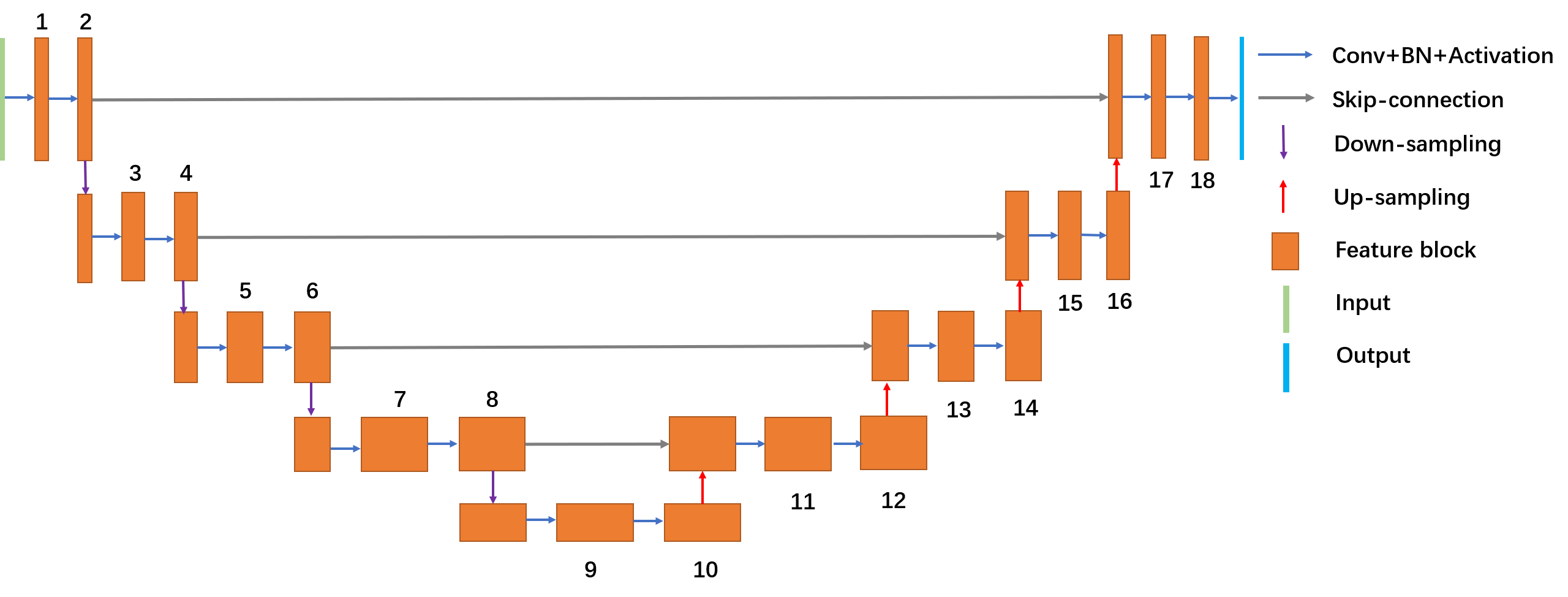}
\caption{Structure of UNet. The numbers represent the number of the layer.}
\label{unet}
\end{figure}  

\begin{figure}[h]
\centering
\includegraphics[scale=0.03]{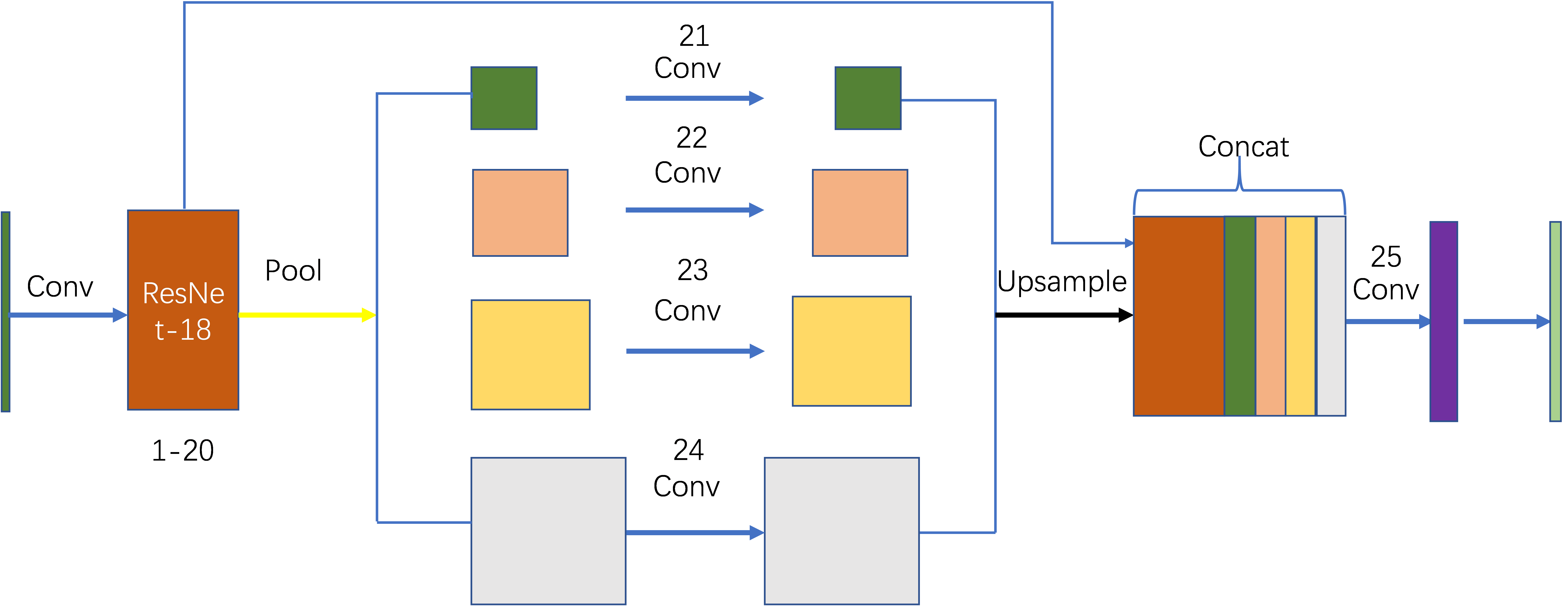}
\caption{Structure of PSPNet. The numbers represent the number of the layer.}
\label{pspnet}
\end{figure}  

The structure of the article is as follows:
In the section of Theory, the BN layer formula and the corresponding formula after the replacement of each parameter are introduced.
	
Experiment is divided into three parts. The first part conducts parameters replacement experiment and proves the validity of the formula in the section Theory. The second part explores the differences between different data sets and different task network parameters. The third part carries on the parameter reuse experiment.

\section{Theory}
Input: Values of $x$ over a mini-batch: $B=\left\{x_{1 \ldots n}\right\}$ ,$x_i = \{x_{i1 \ldots im}\}$,$m$ is the number of channel. Parameters to be learned: $w=\left\{w_{1 \ldots m}\right\},b=\left\{b_{1 \ldots m}\right\}$.   

Output: $\left\{y_{ij}=\mathrm{B} \mathrm{N}_{w_j, b_j}\left(x_{ij}\right)\right\}$.  
$$
\mu_{j} \leftarrow \frac{1}{m} \sum_{i=1}^{m} x_{ij}\  \  \  //mini-batch\   mean  \\
$$
$$
\sigma_{j}^{2} \leftarrow \frac{1}{m} \sum_{i=1}^{m}\left(x_{ij}-\mu_{j}\right)^{2} \  \  \   //mini-batch\   variance  \\ 
$$
$$
\widehat{x}_{i} \leftarrow \frac{x_{ij}-\mu_{j}}{\sqrt{\sigma_{j}^{2}+\epsilon}}\  \  \  //normalize \\ 
$$
$$
y_{ij} \leftarrow w_j \widehat{x}_{ij}+b_j \equiv \mathrm{B} \mathrm{N}_{w_j, b_j}\left(x_{ij}\right)  \  \  \  //scale\   and \   shift
$$ 

The Running mean,Running variance,Weight,Bias(in pytorch framework) in BN layer represent $\mu,\sigma,w,b$. They are called as RM,RV,RW,RB in this paper. $\epsilon$ is a very small value.Change the parameters of BN layer RM, RV, RW, RB, BN formula can be written as follows. 

{\bfseries RM}
$$
\begin{array}{l}
\hat{x}_{ij}=\frac{x_{ij}-\left(\mu_{j} \pm \Delta \mu_{j}\right)}{\sqrt{\sigma_{j}^{2}+\varepsilon}}\left(\Delta \mu_{j}=\mu_{j}-\mu_{j}^{\prime}\right) \\
\hat{x}_{ij}=\frac{x_{i}-\mu_{j}}{\sqrt{\sigma_{j}^{2}+\varepsilon}} \pm \frac{\Delta \mu_{j}}{\sqrt{\sigma_{j}^{2}+\varepsilon}} \\
y_{ij}=w_j \frac{x_{ij}-\mu_{j}}{\sqrt{\sigma_{j}^{2}+\varepsilon}} \pm w_j \frac{\Delta \mu_{j}}{\sqrt{\sigma_{j}^{2}+\varepsilon}}+b_j
\end{array}
$$

{\bfseries RV}
$$
\begin{array}{l}
\hat{x}_{ij}=\frac{x_{ij}-\mu_{j}}{\sqrt{\sigma_{j}^{2} \pm \Delta\sigma_j^{2}+\varepsilon}} \quad\left(\Delta \sigma_{j}^{2}=\sigma_{j}^{2}-\sigma_{j}^{\prime 2}\right) \\
\hat{x}_{ij}=\frac{x_{ij}-\mu_{j}}{\sqrt{\sigma_{j}^{2}+\varepsilon}} \cdot \frac{\sqrt{\sigma_{j}^{2}+\varepsilon}}{\sqrt{\sigma_{j}^{2} \pm \Delta \sigma_{j}^{2}+\varepsilon}} \\
y_{ij}=w_j \frac{x_{ij}-\mu_{j}}{\sqrt{\sigma_{j}^{2}+\varepsilon}} \frac{\sqrt{\sigma_{j}^{2}+\varepsilon}}{\sqrt{\sigma_{j}^{2} \pm \Delta \sigma_{j}^{2}+\varepsilon}}+b_j
\end{array}
$$

{\bfseries RW}
$$
y_{ij}=\alpha_j w_j \hat{x}_{ij}+b_j \quad \quad\left(\alpha_j=\frac{w_j^{\prime}}{w_j}\right)
$$

{\bfseries RB}
$$
y_{ij}=w_j \hat{x}_{ij}+b_j \pm \Delta b_j \quad\left(\Delta b_j=b_j-b_j^\prime\right)
$$

In the above formula $\mu_j^\prime,\sigma_{j}^{\prime 2},w_j^{\prime},b_j^\prime$ represent RM,RV,RW,RB of auto-encoder task at BN layer,and the others represent the parameters from segmentation task.

\section{Data}
Three data sets were used in this experiment, namely ACDC\cite{ACDC},T1\cite{T1} and RVSC\cite{RVSC}.  

ACDC (Reference) data set is used to segment the left ventricle, right ventricle and myocardium. There are 100 people in the data set, and each person has 10-20 images. The data set was randomly divided into two parts, a train set and a validation set. Each containing 50 people.T1 data set was used to segment the myocardium. There are 210 people in total, with 55 images per person. The first 60 people were taken as the train set, and the last 30 people were taken as the validation set. There are three groups of RVSC: TrainingSet, Test1Set and Test2Set, each with 16 sample. In this experiment, only TrainingSet and Test1Set were used, with 242 and 262 images respectively.

\section{Experiment}
\subsection{1.Parameter replacement for each layer of the network}
UNet and PSPNet networks were used to train image auto-encoder and image segmentation tasks on the ACDC data set. Since the number of channels in the segmentation result is 4, to ensure the correspondence of parameters, the image auto-encoder task also outputs 4 channels, each of which is the original image. The parameters of the corresponding layer of image segmentation were replaced layer by layer by image auto-encoder task parameters, and the corresponding Dice value was calculated on the validation set.

\begin{table}[H]
    \caption{ACDC:Result of UNet}
    \centering
    \begin{tabular}{|l|l|l|l|l}
        class & class-0 & class-1 & class-2 & class-3 \\ 
        Dice & 1.00 & 0.81 & 0.83 & 0.92 \\
    \end{tabular}
\end{table}

\begin{table}[H]
\caption{ACDC:Result of PSPNet}
    \centering
    \begin{tabular}{|l|l|l|l|l|}
        PSPNet-b & class-0 & class-1 & class-2 & class-3 \\ 
        yes & 1 & 0.86 & 0.86 & 0.93 \\ 
        no & 1 & 0.86 & 0.86 & 0.92 \\ 
    \end{tabular}
\end{table}

Since the pre-trained parameters of ResNet-18 were loaded during the training, but there is no bias parameters in the convolutional layers of ResNet-18, so only part of the convolutional layers contained bias parameters. In Table 2, yes, no means whether the convolutional layers of the models contain bias.

\subsubsection{1.1 UNet}

\paragraph{1.1.1 Result of UNet}~{}

{\bfseries Parameter replacement results in BN layers:} 

The parameters of the BN layer of image segmentation were replaced by the parameters of the corresponding layer of image auto-encoder task layer by layer, and the corresponding Dice values were obtained on the validation set, and the results are as follows:  

\begin{figure}[h]
\centering
\includegraphics[scale=0.05]{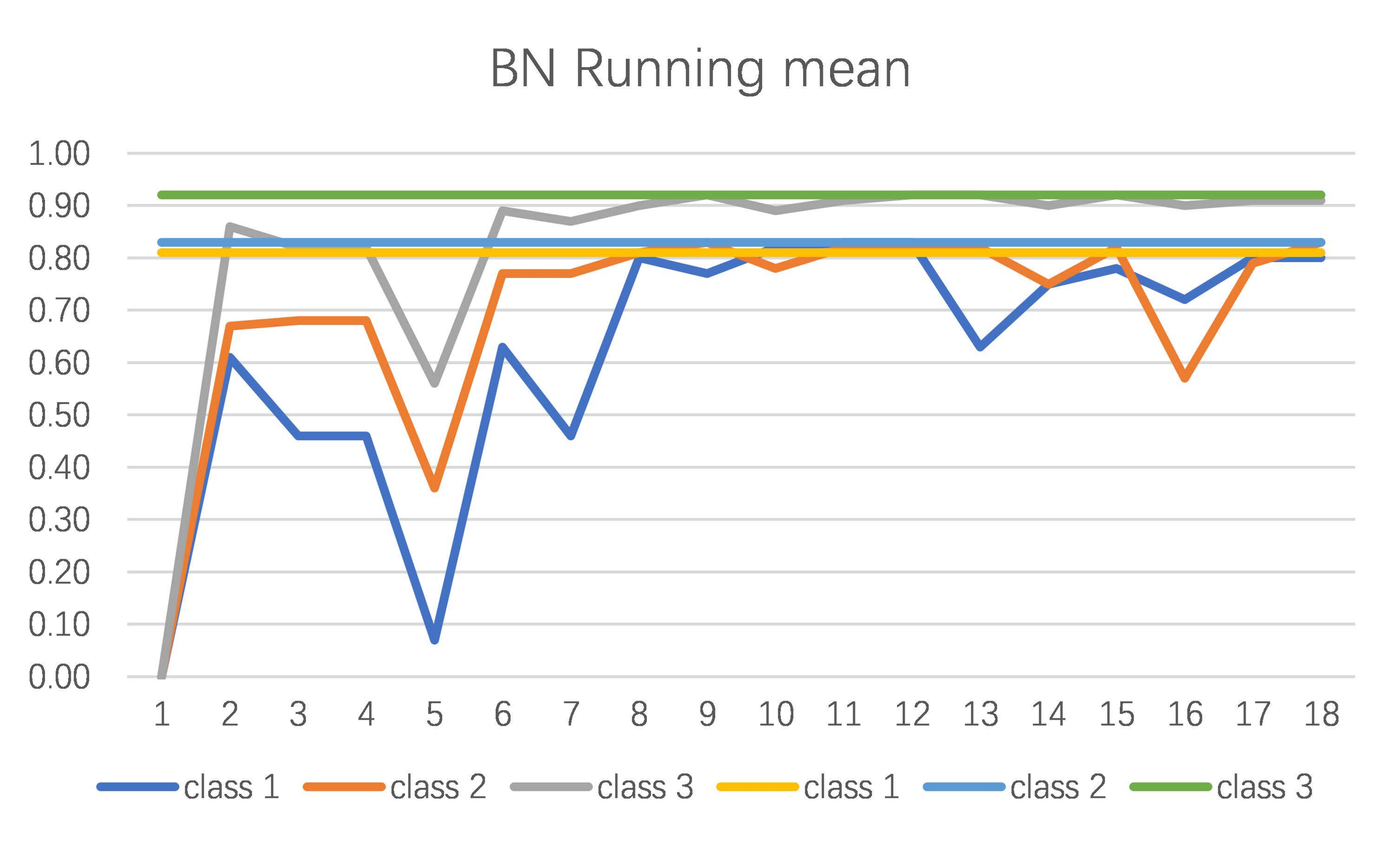}
\includegraphics[scale=0.05]{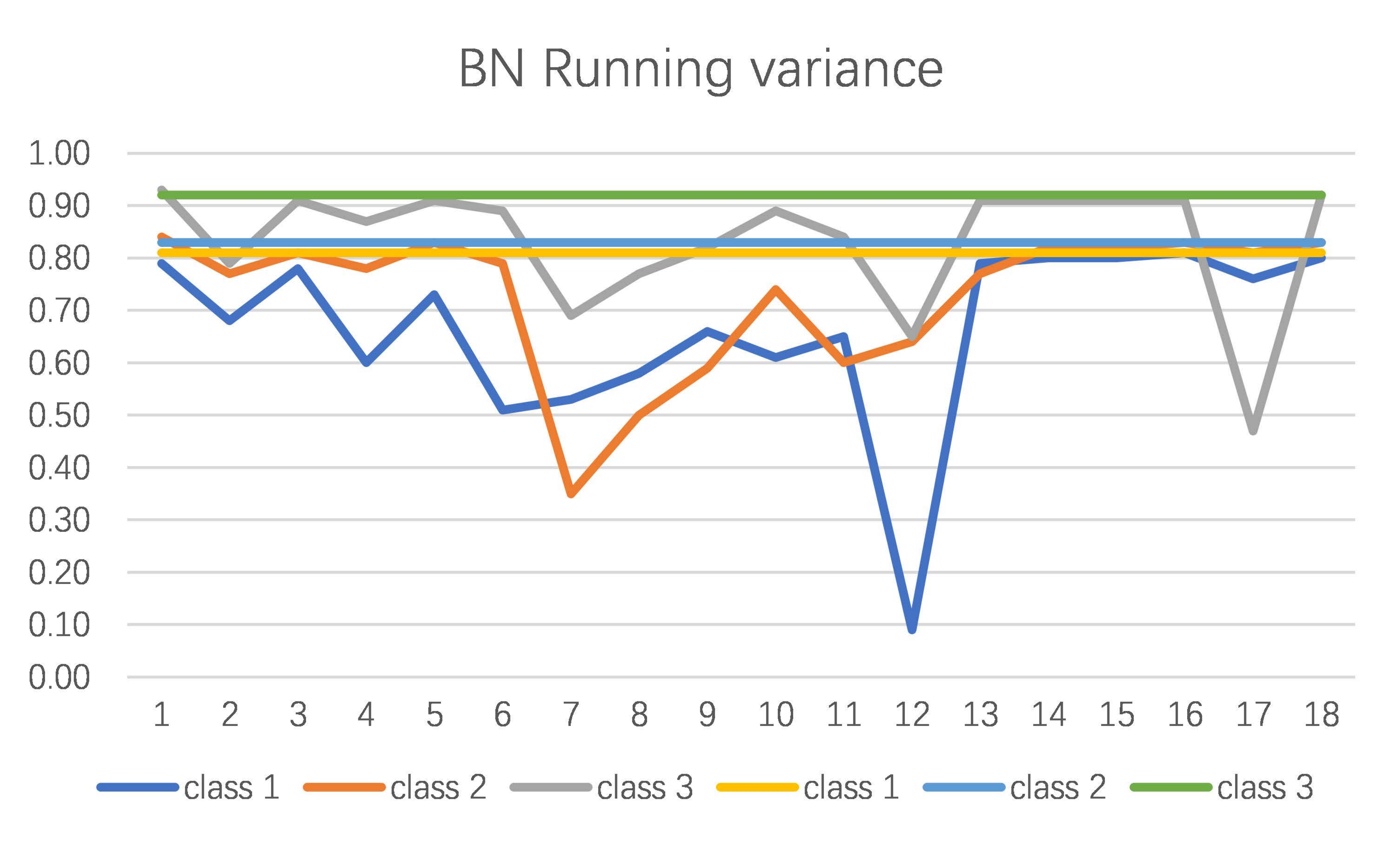}

\includegraphics[scale=0.05]{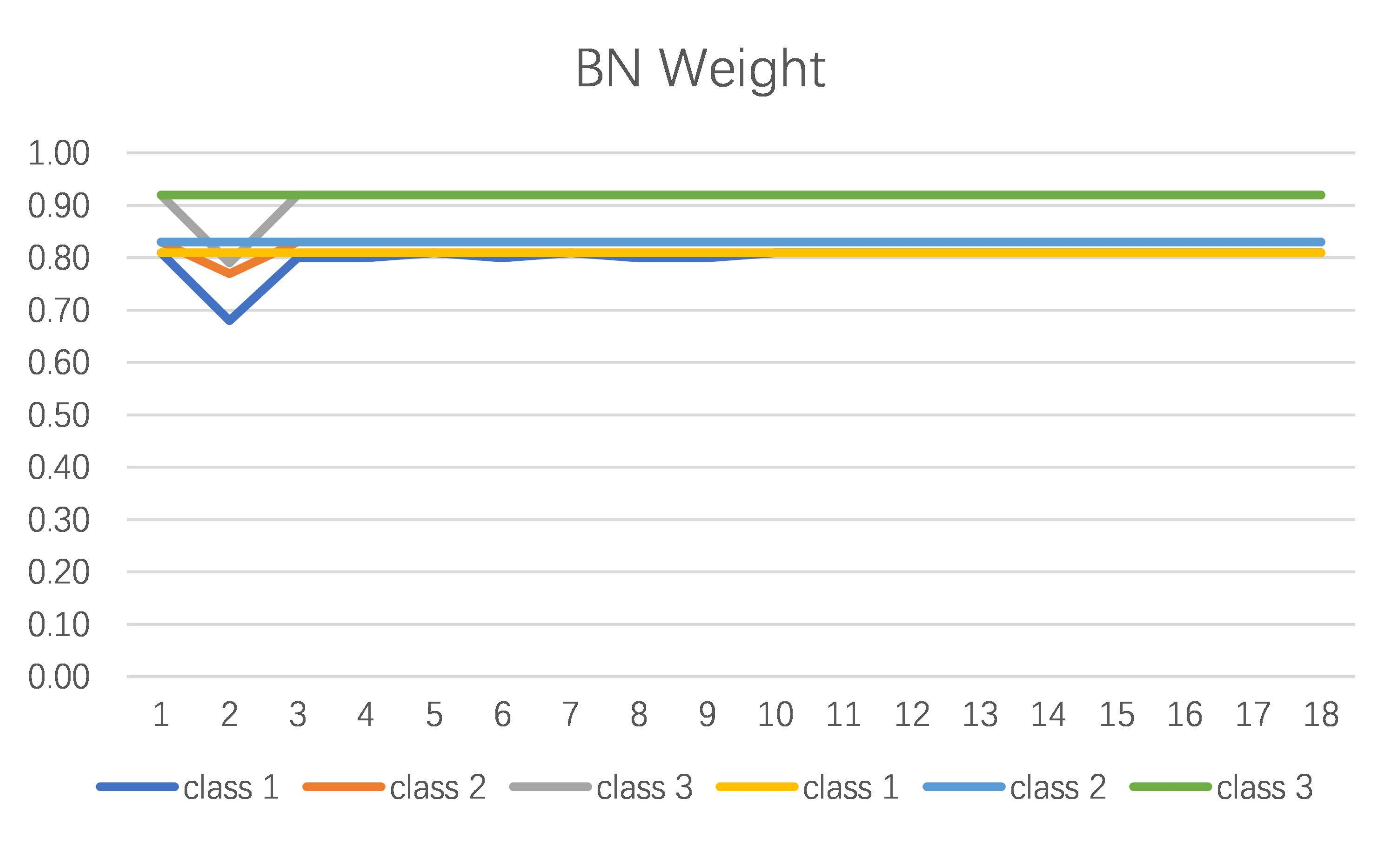}
\includegraphics[scale=0.05]{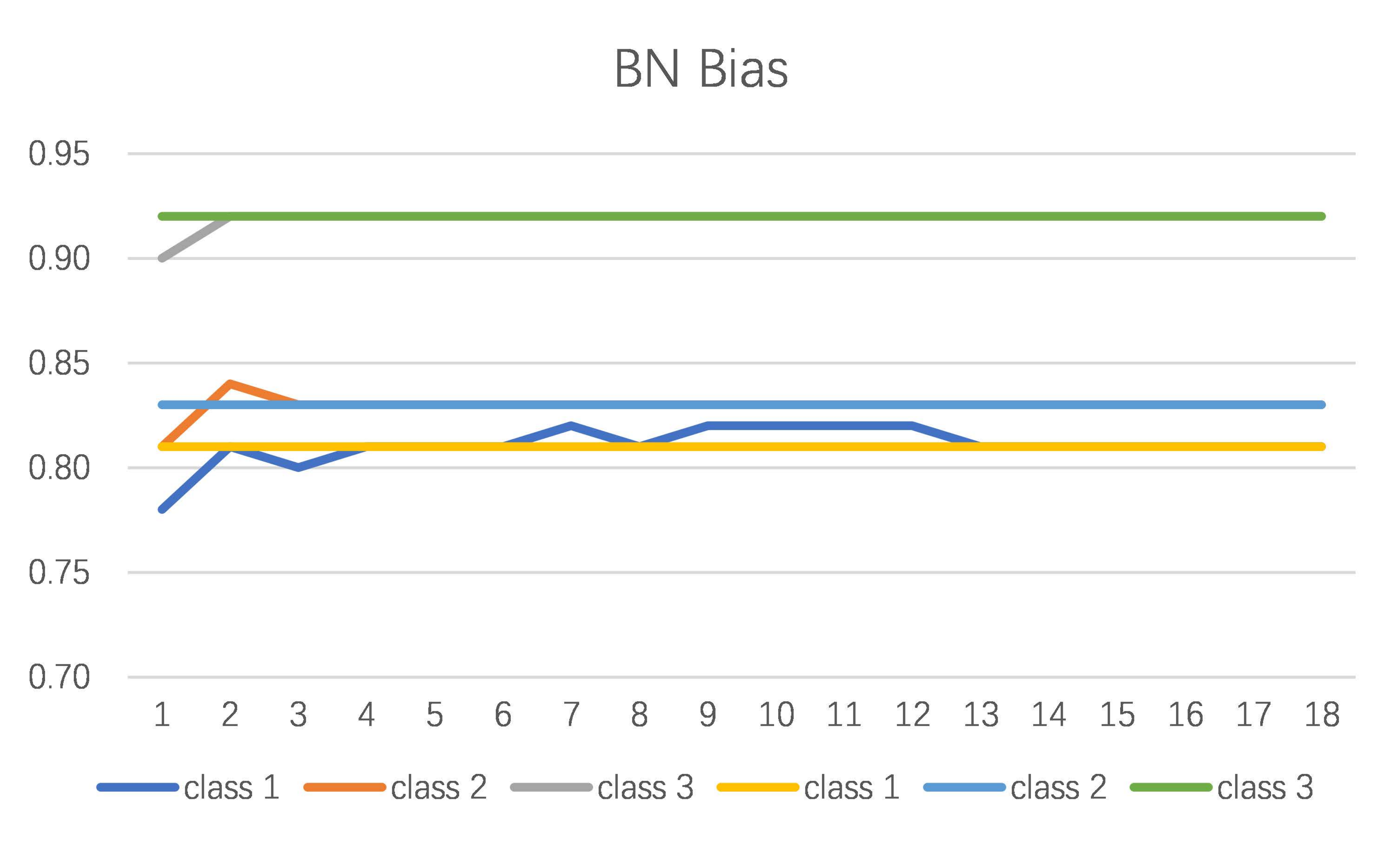}

\includegraphics[scale=0.05]{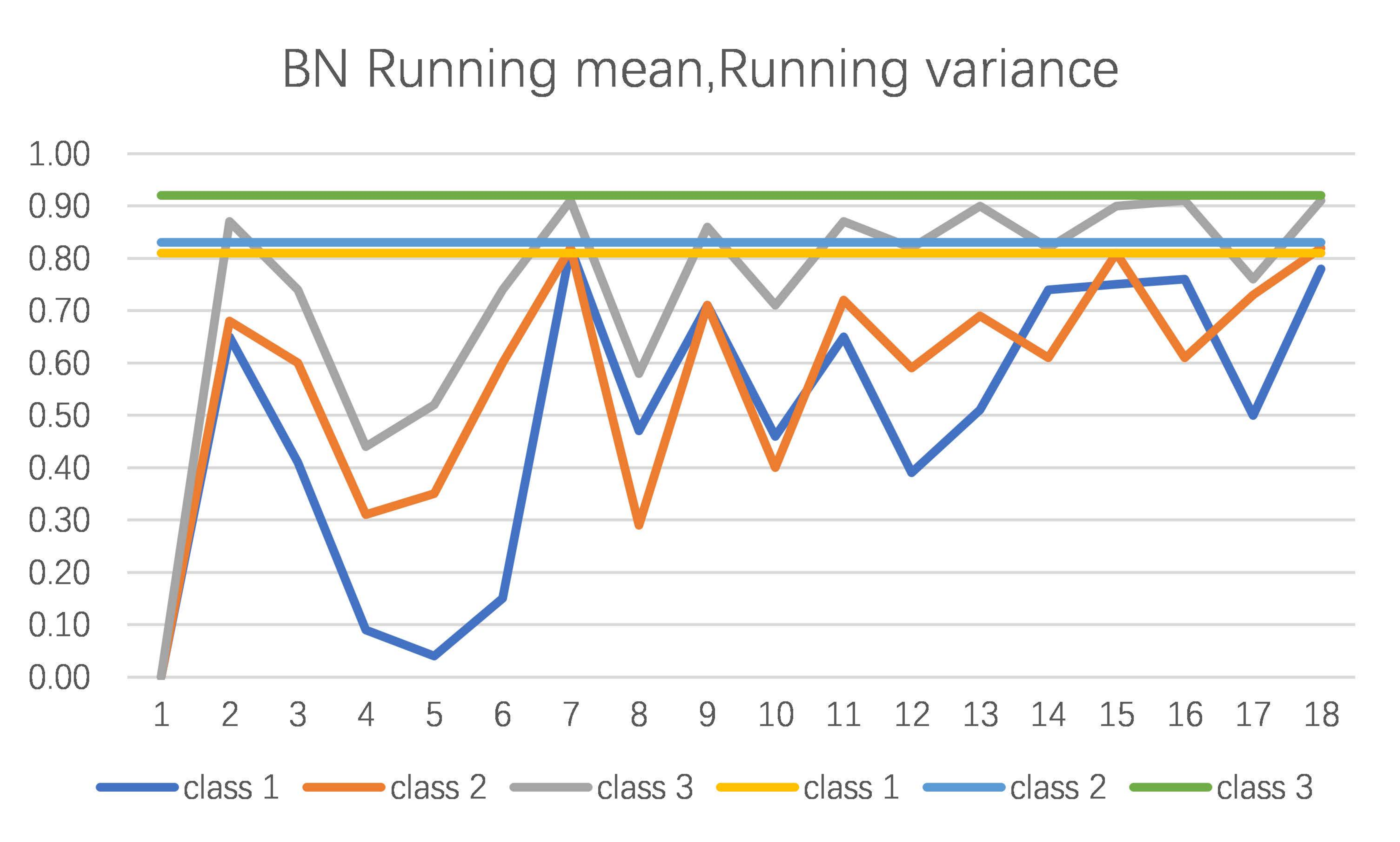}
\includegraphics[scale=0.05]{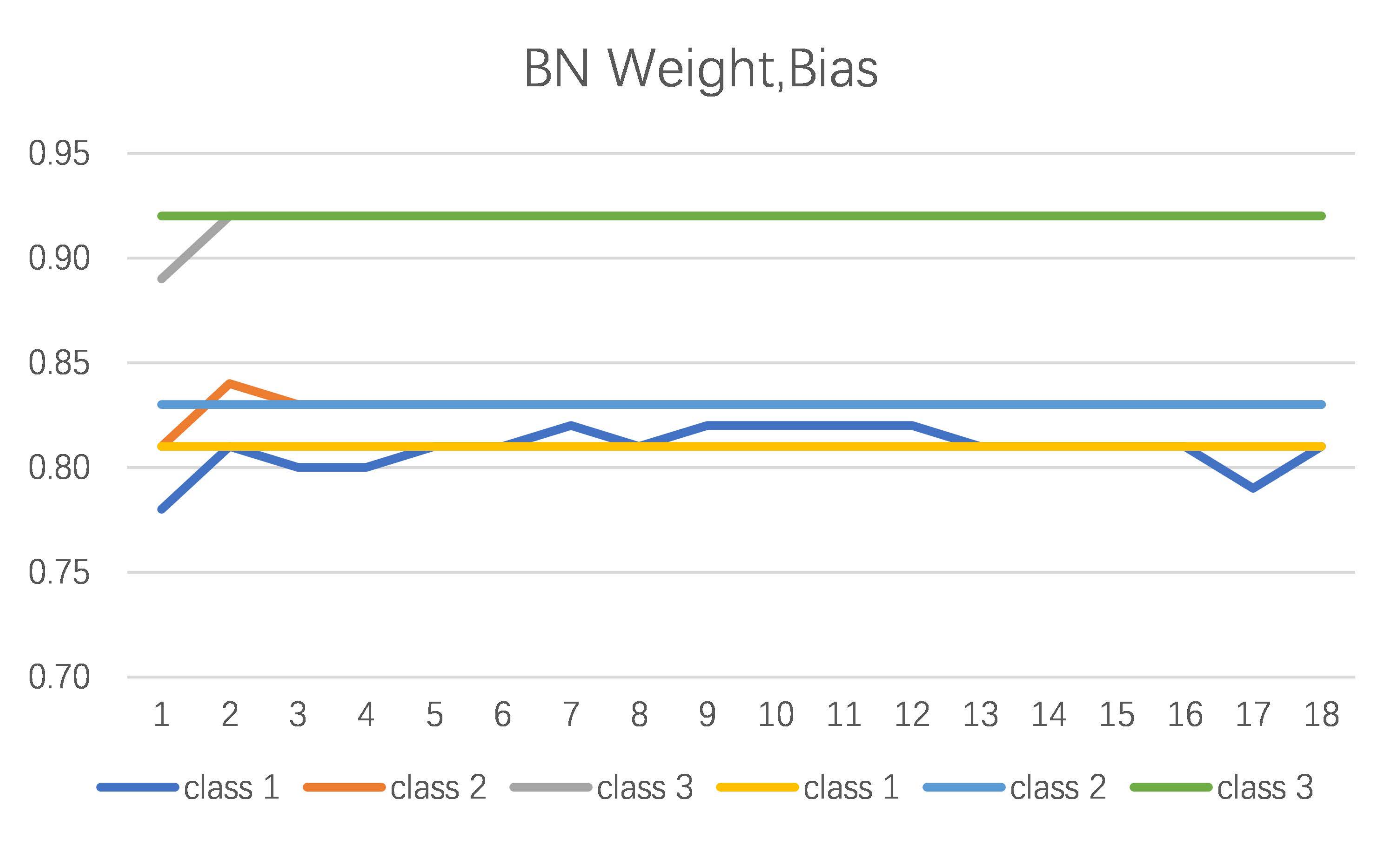}
\caption{UNet : The six images show the Dice value when the titled parameters were replaced,
and the abscissa represents the layer to be replaced ( 1-18 corresponds to the BN layers of the
network from front to back). The ordinate represents the corresponding Dice value after parameter
replacement. class 1: dark blue,yellow;class 2: orange,baby blue;class 3: gray, green.(The first position of the classes represent the result of parameter replaced, and the second represent the baseline)
}
\label{unet-bn}
\end{figure} 

{\bfseries Parameter replacement results in Conv layers:} 
\begin{figure}[H]
\centering
\includegraphics[scale=0.05]{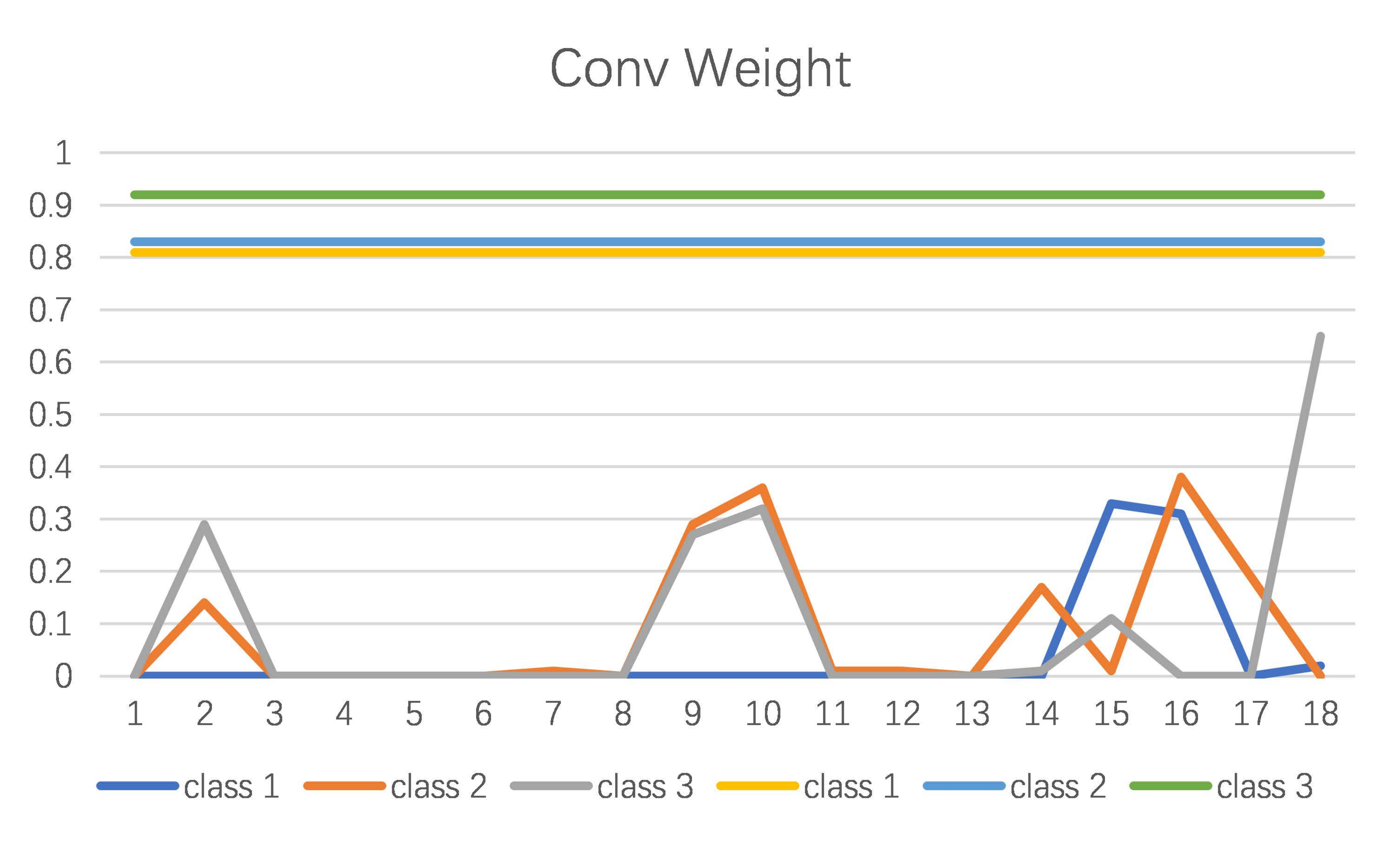}
\includegraphics[scale=0.05]{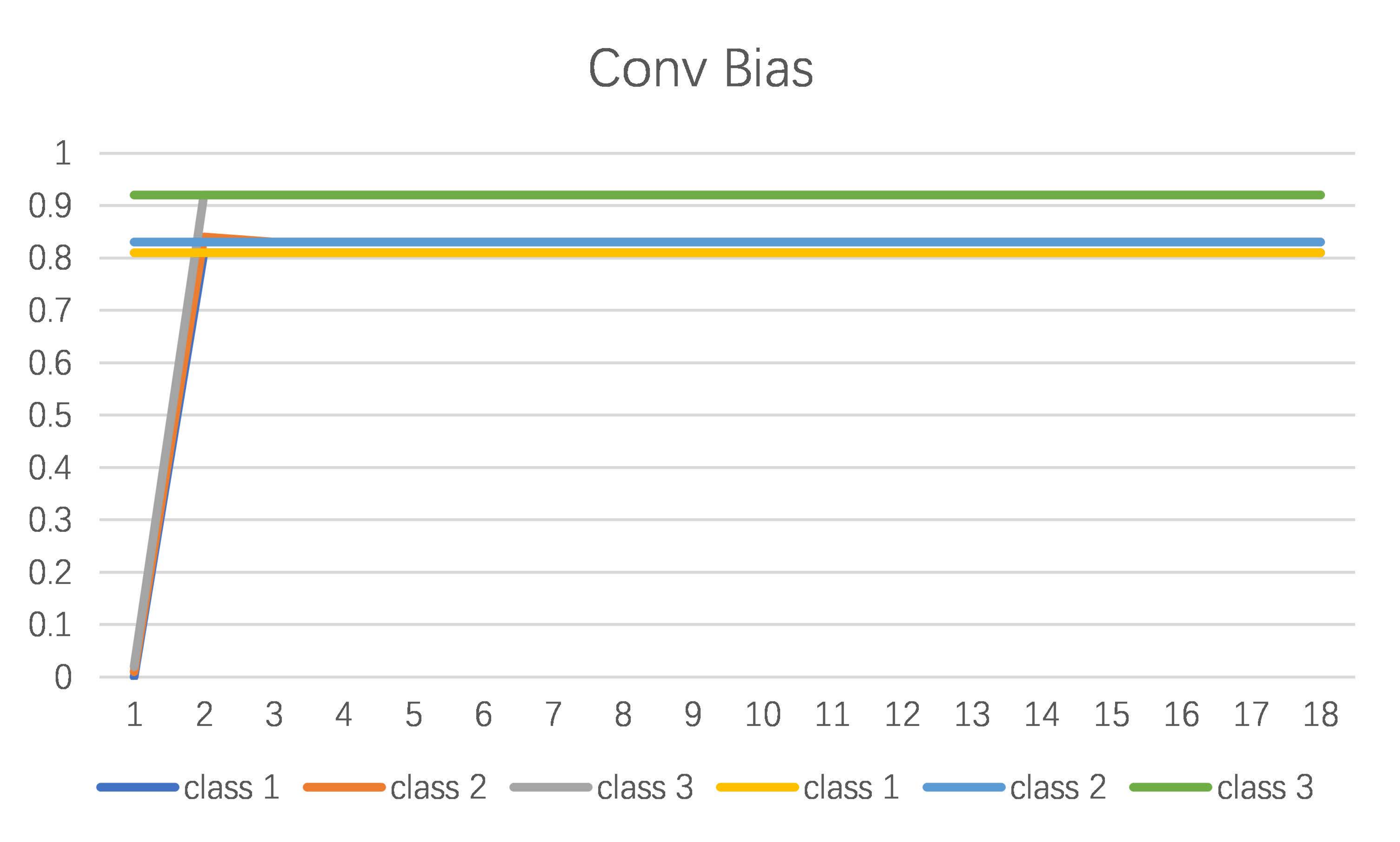}
\includegraphics[scale=0.05]{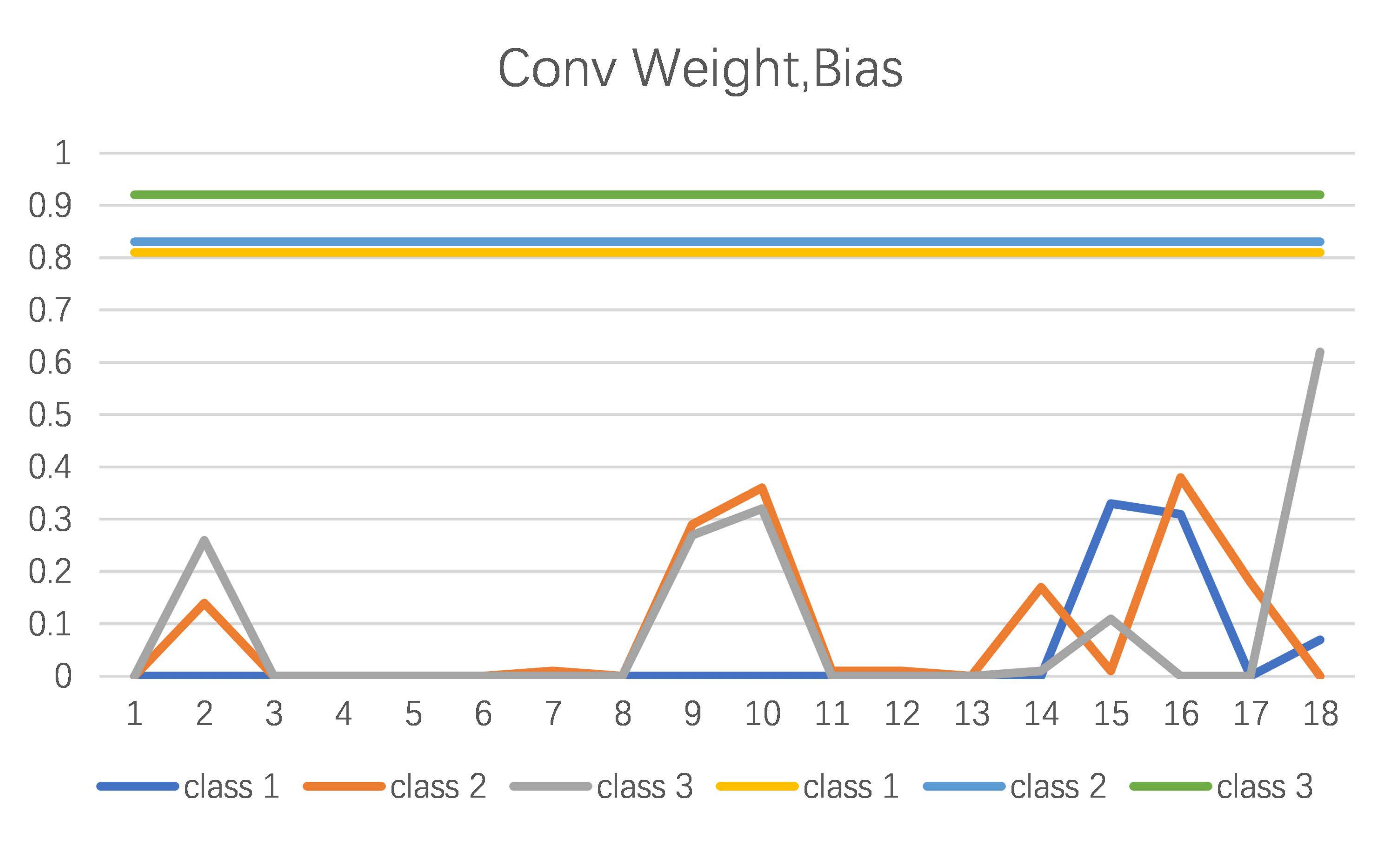}

\caption{UNet: Weight and Bias in the figure correspond to kernel and bias of the Conv layer
respectively. The title in the figure represents the parameters of the Conv layer to be replaced.
The abscissa represents the layer to be replaced(1-18 corresponds to the Conv layer of the network
from front to back). class 1: dark blue,yellow;class 2: orange,baby blue;class 3: gray, green.( The first position of the classes represent the result of parameter replaced, and the second represent the baseline)
}
\label{unet-conv}
\end{figure} 

\paragraph{1.1.2 Explore the reason of UNet result}~{}

{\bfseries Scale and shift of BN Layers:} 

According to the results shown in Fig3, the network segmentation results are influenced more by RM, RV than RW and RB. So let's assume that $w_j\frac{\mathrm{\Delta}\mu_j}{\sqrt{\sigma_j^2+\varepsilon}}>>\mathrm{\Delta\ b_j},\ \frac{\sqrt{\sigma_j^2+\varepsilon}}{\sqrt{\sigma_j^2\pm\mathrm{\Delta}\sigma_j^2+\varepsilon}}>>\alpha_j$.

Plot the figure of $\frac{1}{m}\sum_{j=1}^{m}w_j\frac{\mathrm{\Delta}\mu_j}{\sqrt{\sigma_j^2+\varepsilon}}(\mathrm{\Delta}\mu_j = \sqrt{(\mu_j-\mu_j^\prime)^2}),\frac{1}{m}\sum_{j=1}^{m}\Delta b_j(\Delta b_j=\sqrt{(b_j-b_j^\prime)^2} ),\frac{1}{m}\sum_{j=1}^{m}\frac{\sqrt{\sigma_{j}^{2}+\varepsilon}}{\sqrt{\sigma_{j}^{2} \pm \Delta \sigma_{j}^{2}+\varepsilon}},\frac{1}{m}\sum_{j=1}^{m}\alpha_j  $.

\begin{figure}[H]
\centering
\includegraphics[scale=0.25]{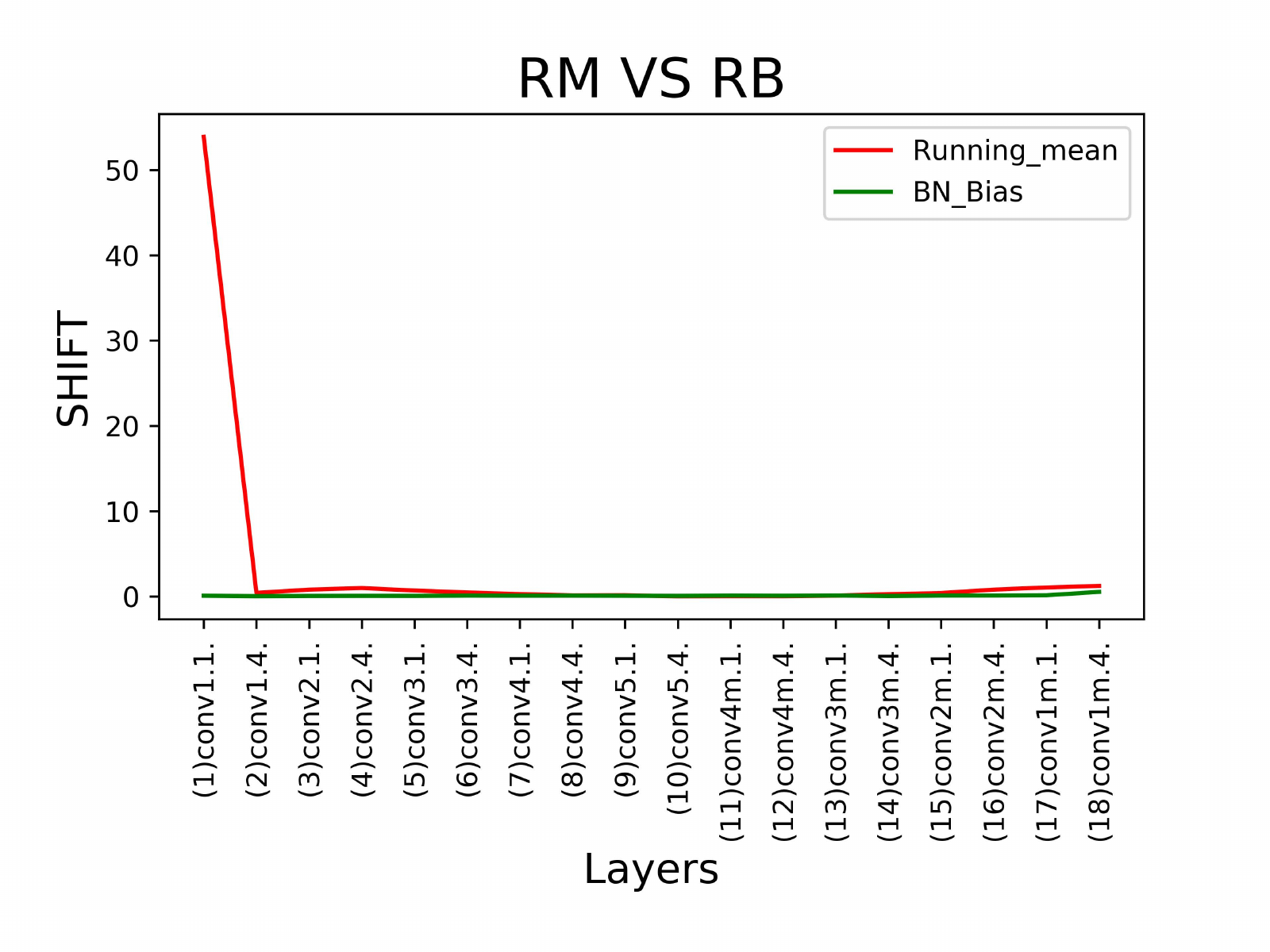}
\includegraphics[scale=0.25]{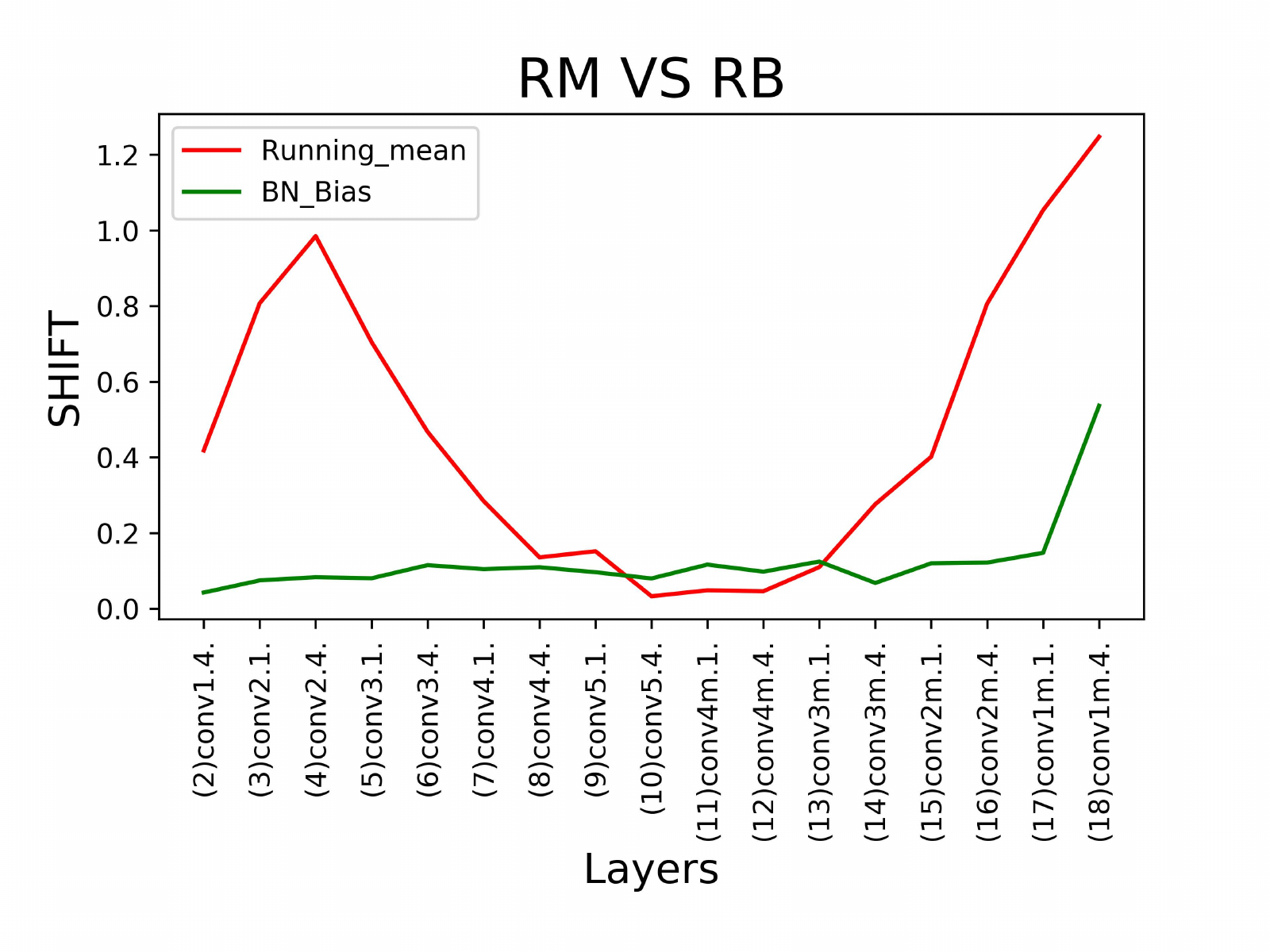}

\includegraphics[scale=0.25]{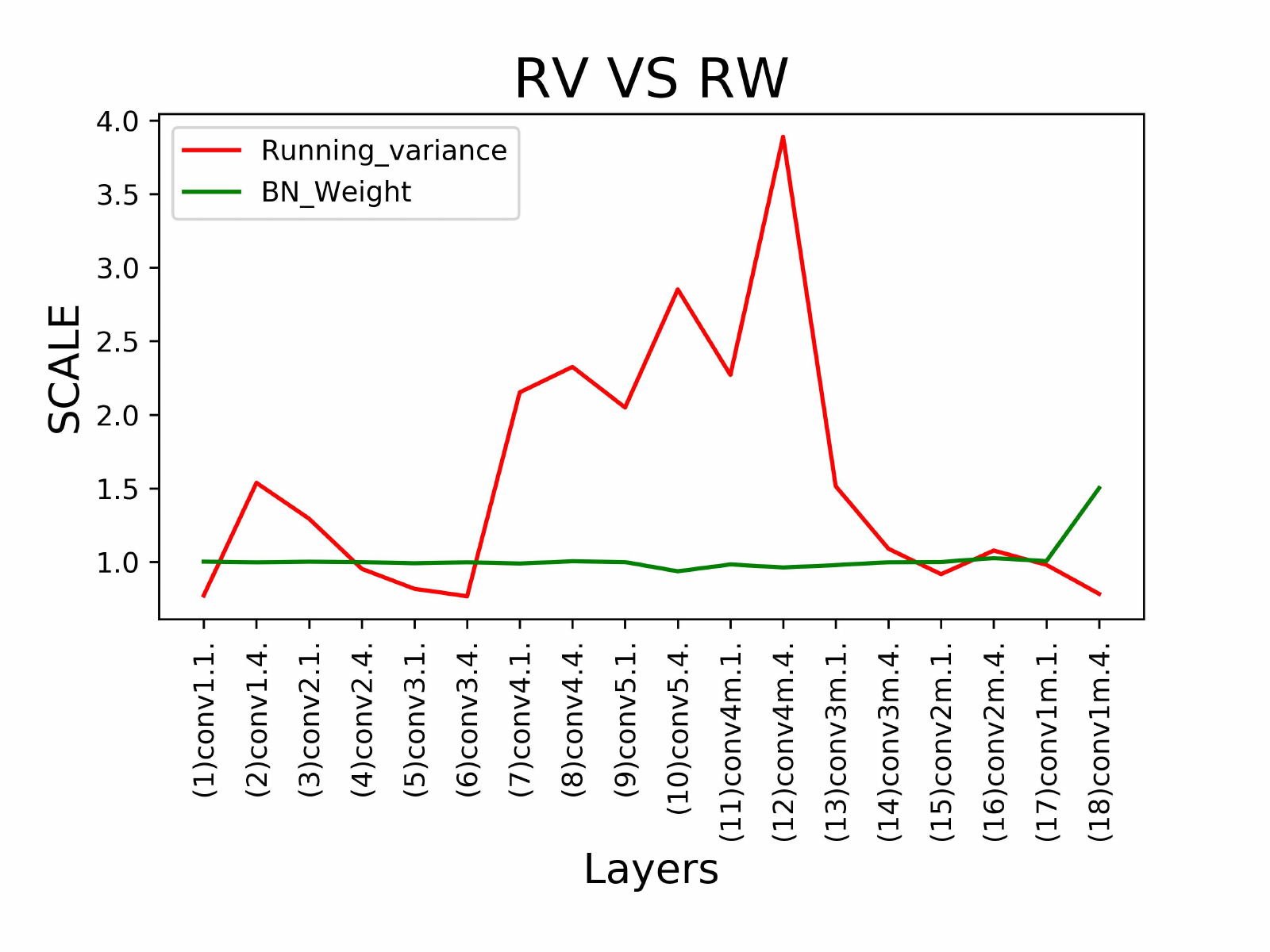}

\caption{UNet:Scale and shift in BN layers.Running Mean:$\frac{1}{m}\sum_{j=1}^{m}w_j\frac{\mathrm{\Delta}\mu_j}{\sqrt{\sigma_j^2+\varepsilon}}(\mathrm{\Delta}\mu_j = \sqrt{(\mu_j-\mu_j^\prime)^2})$;BN Bias:$\frac{1}{m}\sum_{j=1}^{m}\Delta b_j(\Delta b_j=\sqrt{(b_j-b_j^\prime)^2} )$;Running variance:$\frac{1}{m}\sum_{j=1}^{m}\frac{\sqrt{\sigma_{j}^{2}+\varepsilon}}{\sqrt{\sigma_{j}^{2} \pm \Delta \sigma_{j}^{2}+\varepsilon}}$;BN Weight:$\frac{1}{m}\sum_{j=1}^{m}\alpha_j $(The top right image skiped the first layer of RM VS RV).
}
\label{unet-scale-shitf}
\end{figure} 

{\bfseries RMSE of Conv layers:}  

It can be seen from the results in the Figure 4 that the final result of the network in the Conv layers are greatly affected by W and little affected by B. In order to evaluate the differences W and B between the two tasks in Conv layers, RMSE(Root mean square error) images were drawn for each layer. The results are as follows: Figure 6

\begin{figure}[H]
\centering
\includegraphics[scale=0.25]{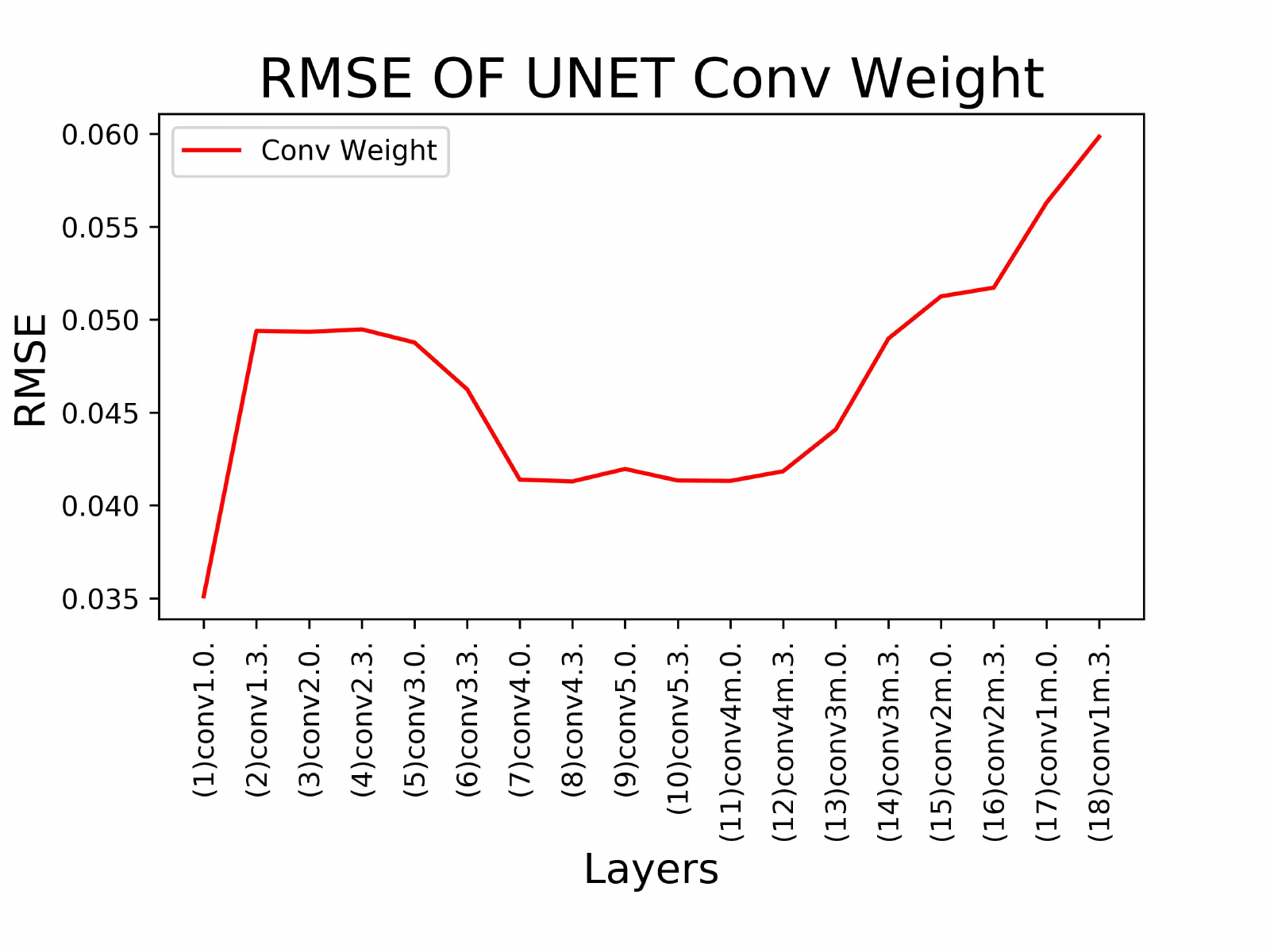}
\includegraphics[scale=0.25]{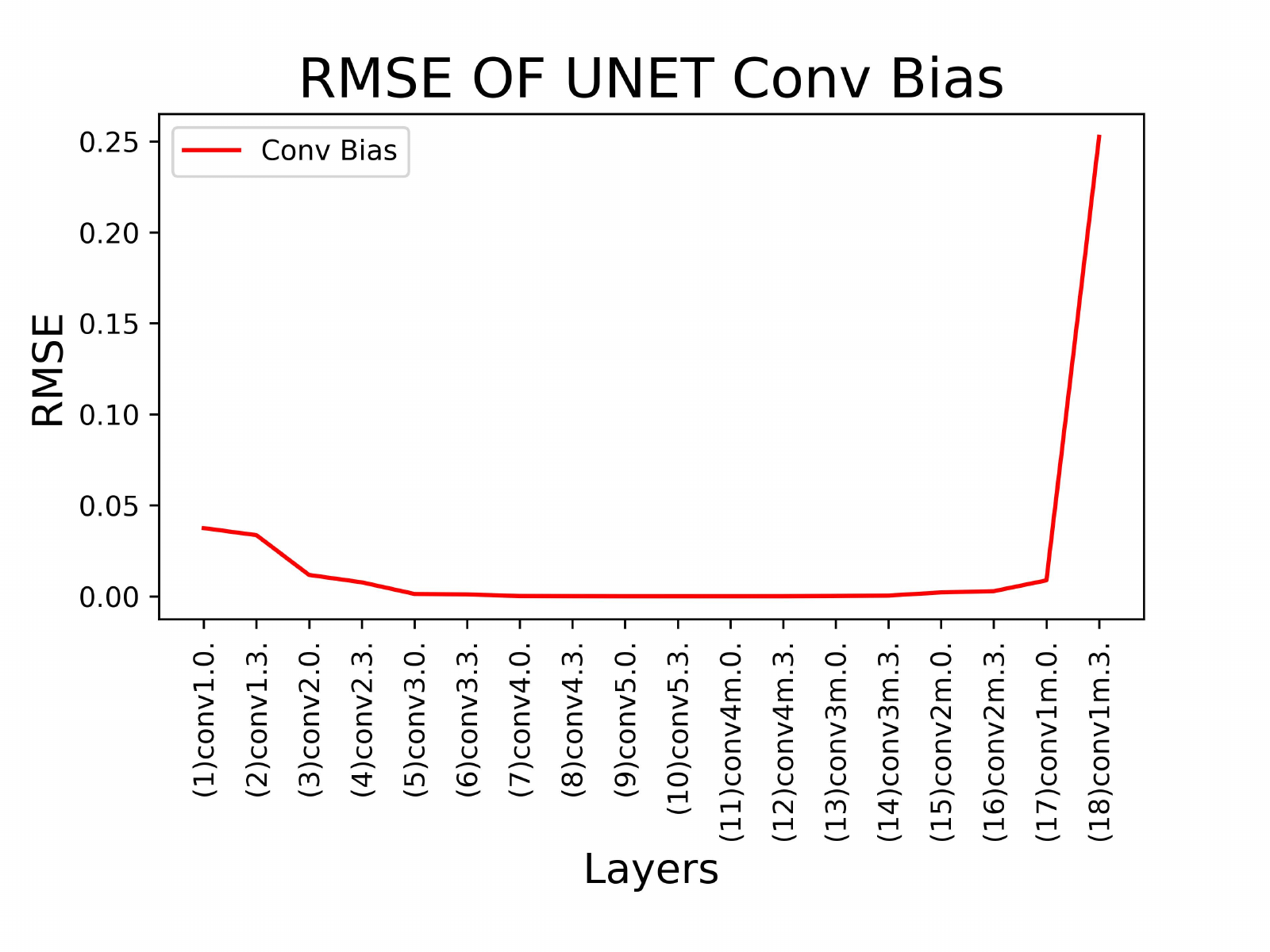}

\caption{UNet:The RMSE of W, B, which are the convolutional layers’ parameters, corresponding
to the image segmentation model and image auto-encoding model.
}
\label{unet-conv-rmse}
\end{figure} 

According to Figure 6 and Figure 4, it can be seen that the network results are very sensitive to the change of W, and even a little change will have a great impact on the results. The results are not sensitive to B.

\subsubsection{1.2 PSPNet}

\paragraph{1.2.1 Result of PSPNet}~{}

{\bfseries Parameter replacement results in BN layers:} 

\begin{figure}[H]
\centering
\includegraphics[scale=0.05]{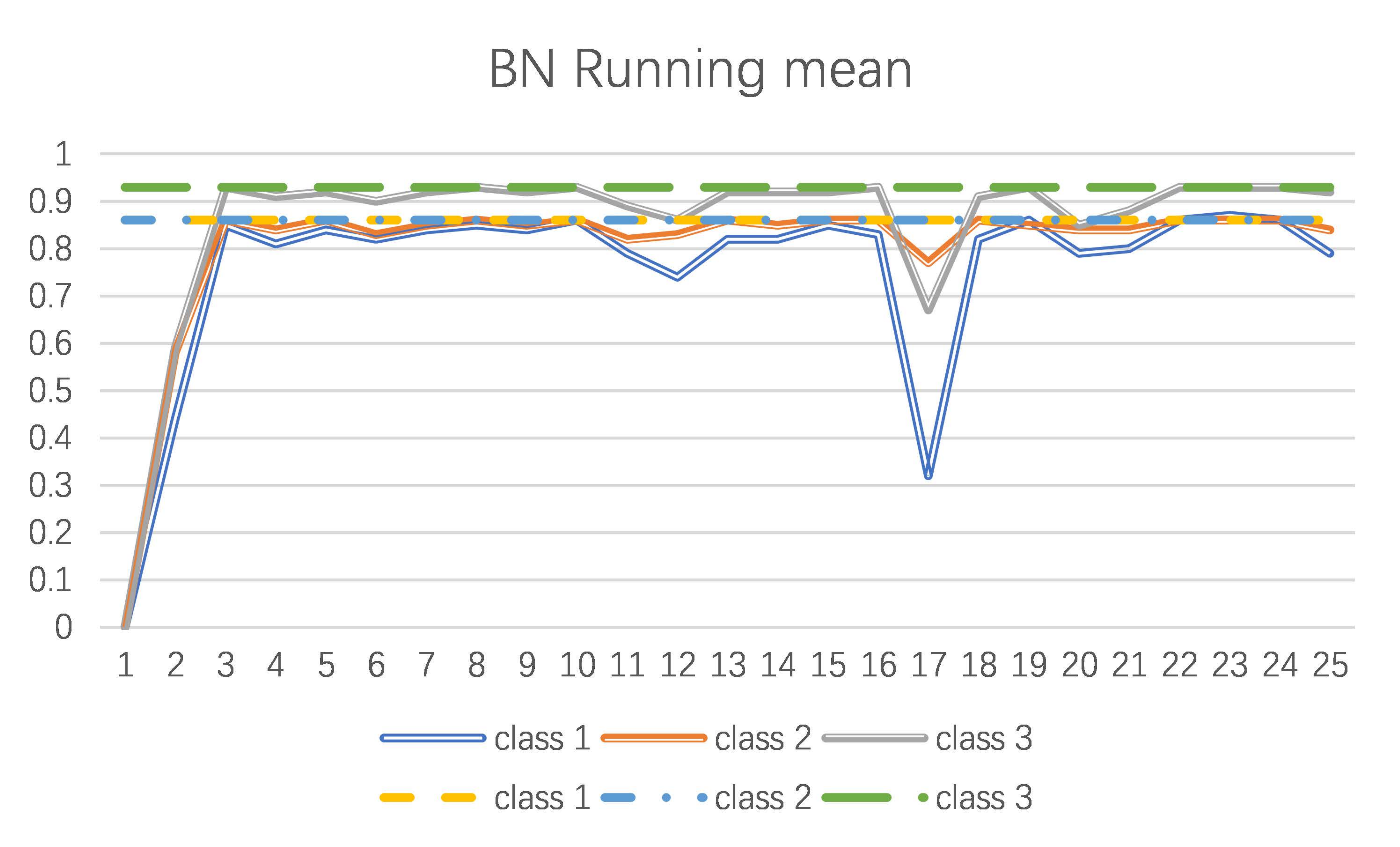}
\includegraphics[scale=0.05]{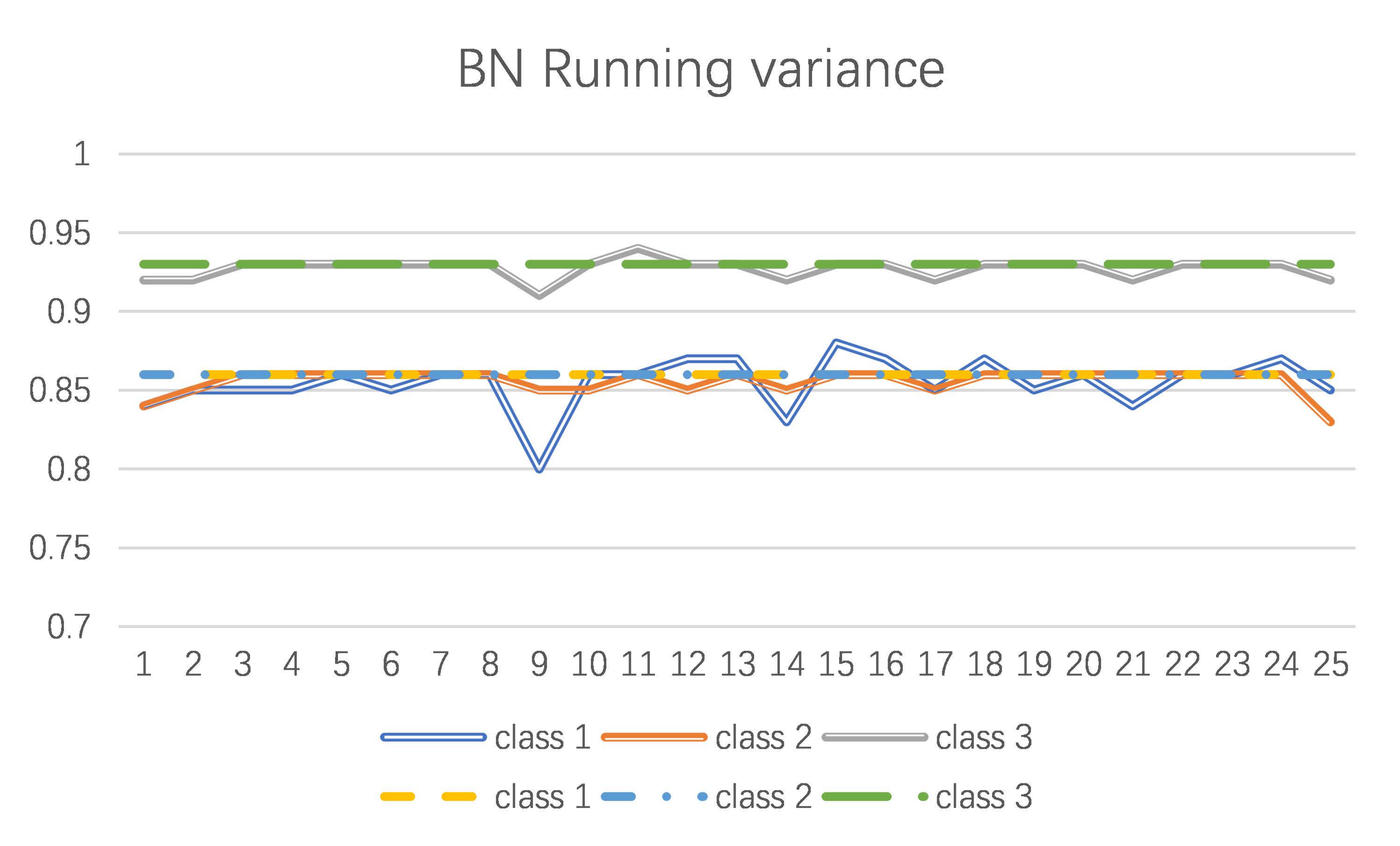}

\includegraphics[scale=0.05]{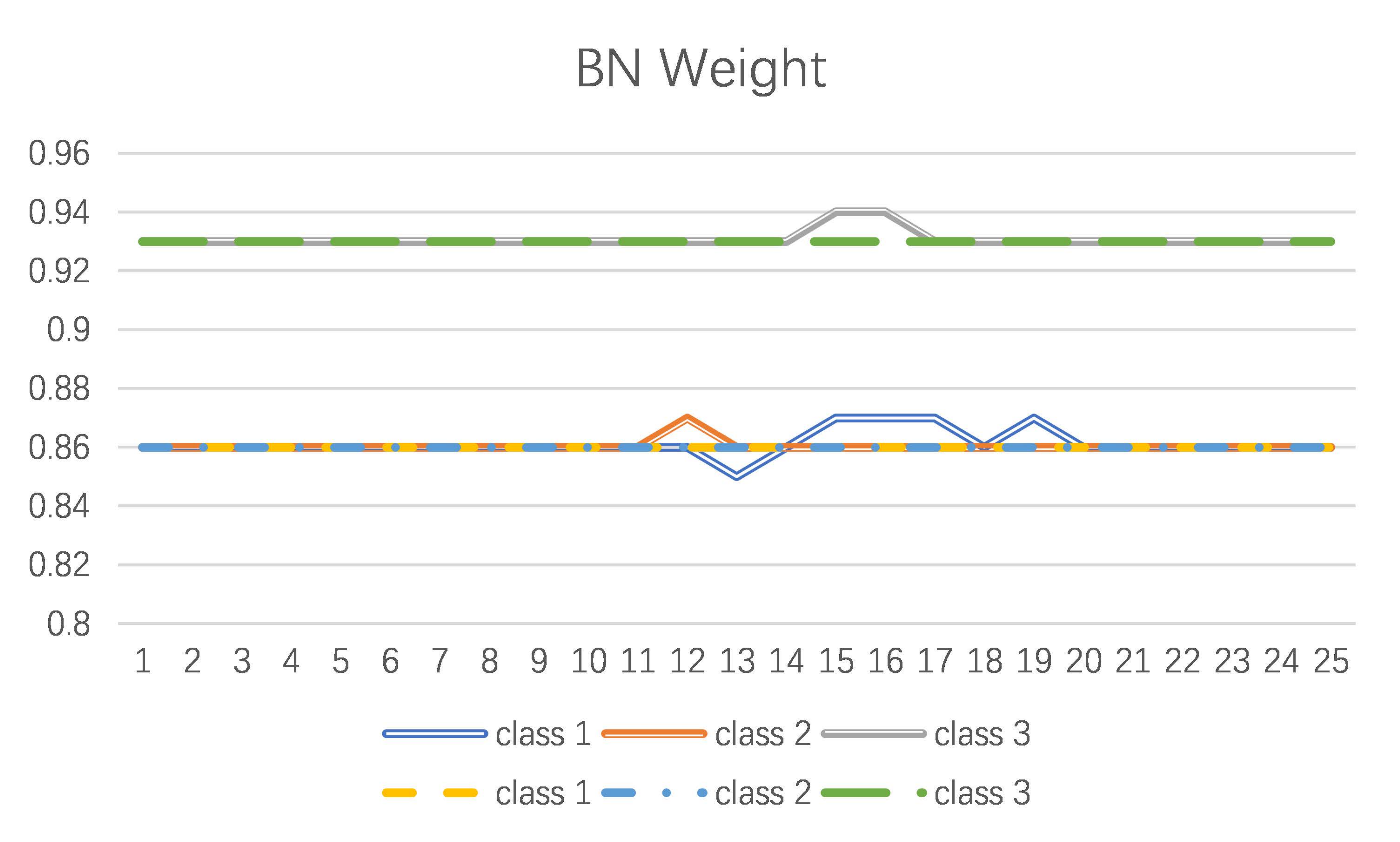}
\includegraphics[scale=0.05]{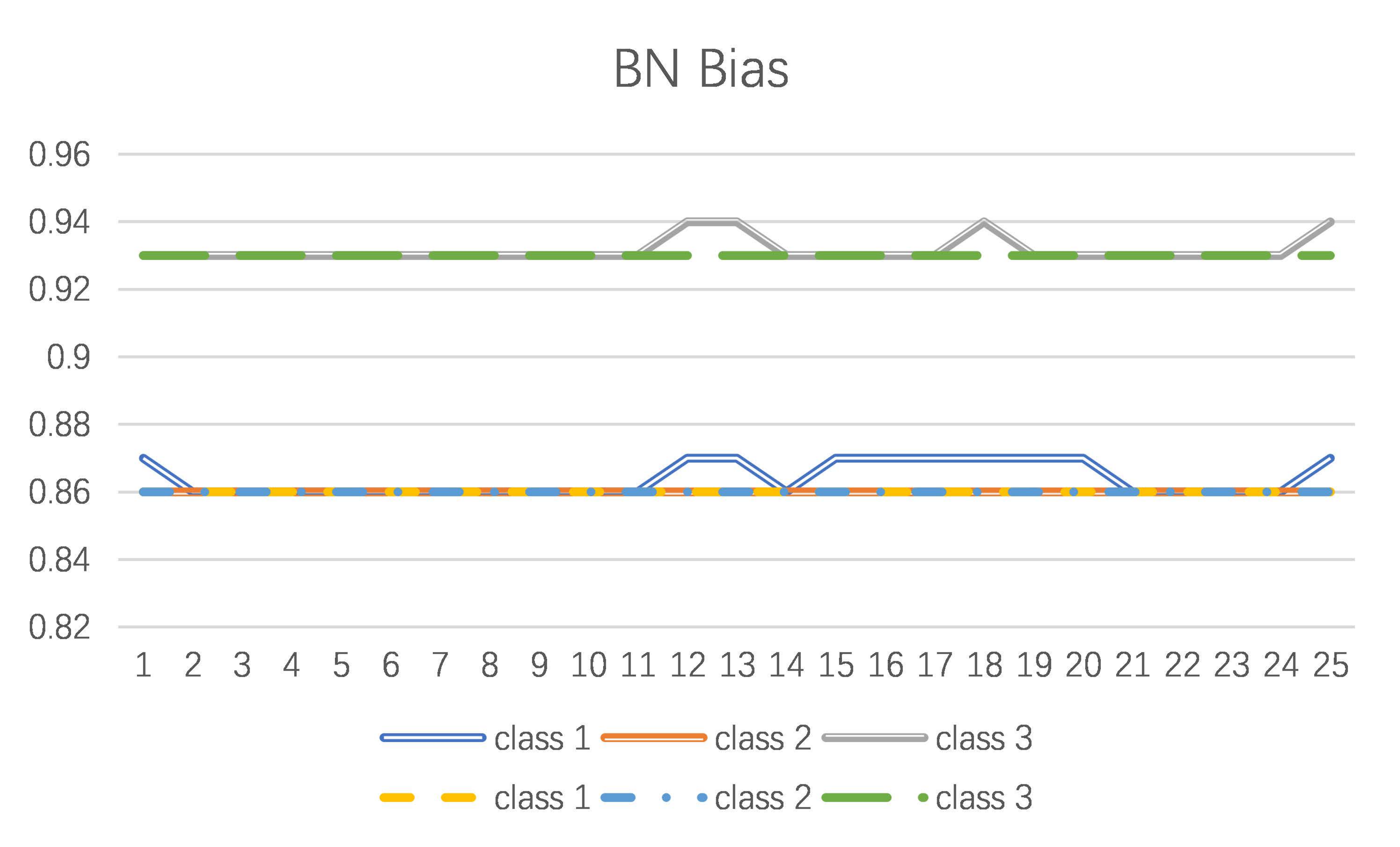}

\caption{PSPNet:The four images show the Dice value when the titled parameters were replaced,
and the abscissa represents the layer to be replaced ( 1-25 corresponds to the BN layers of the
network from front to back). The ordinate represents the corresponding Dice value after parameter
replacement. class 1: dark blue,yellow;class 2: orange,baby blue;class 3: gray, green.(The first position of the classes represent the result of parameter replaced, and the second represent the baseline)}

\label{pspnet-bn}
\end{figure} 

{\bfseries Parameter replacement results in BN layers:} 
\begin{figure}[H]
\centering
\includegraphics[scale=0.05]{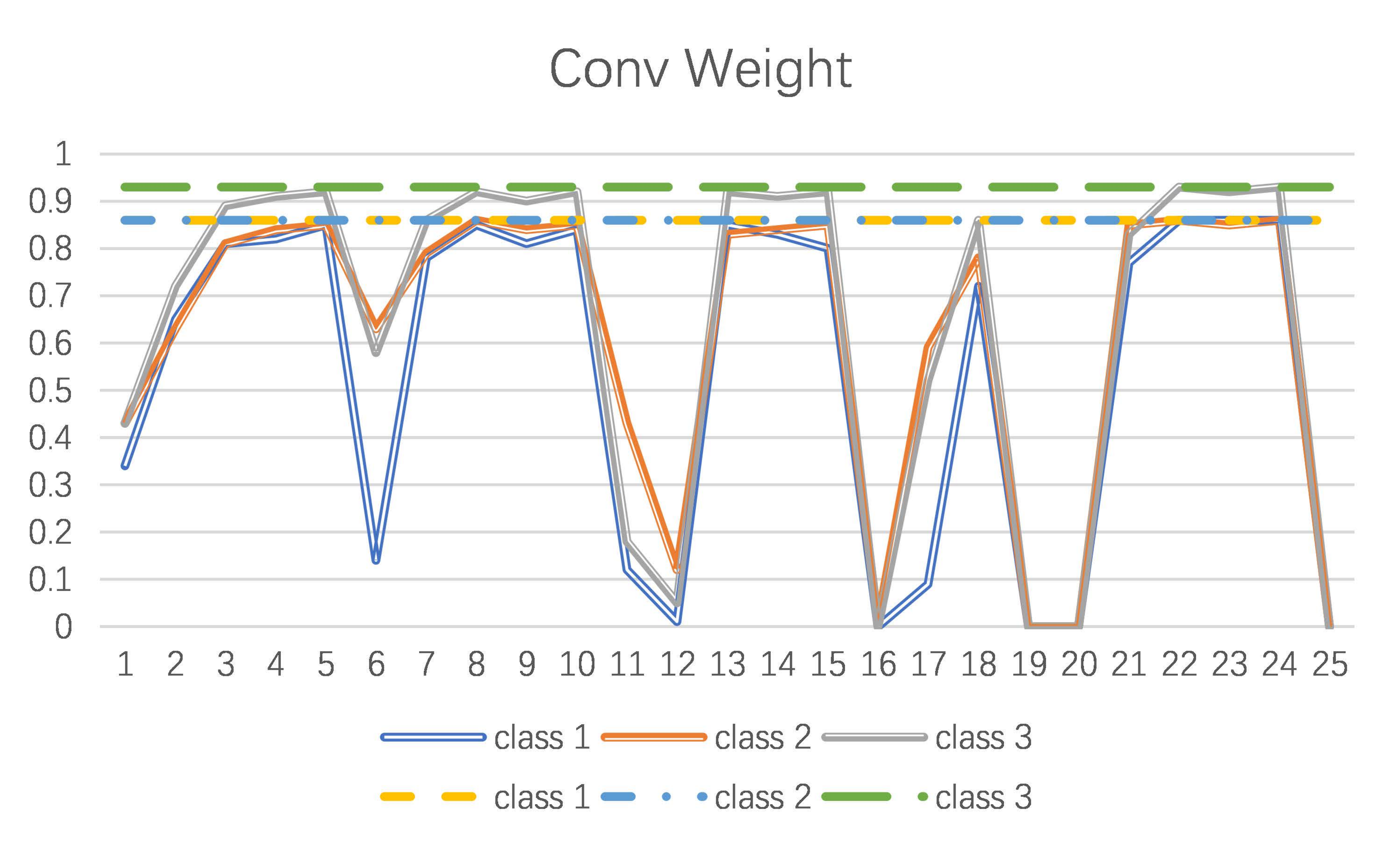}
\includegraphics[scale=0.05]{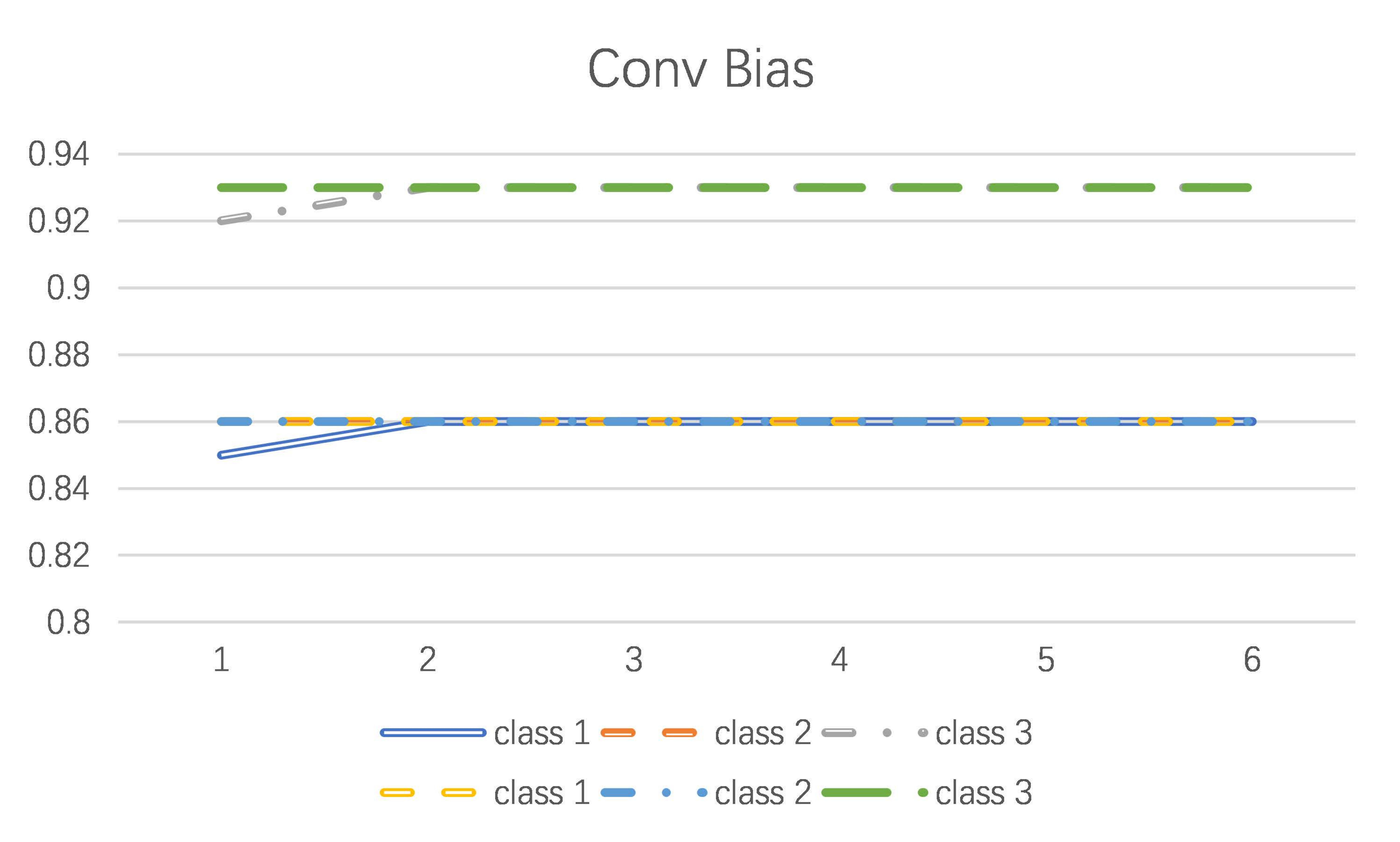}

\caption{PSPNet: Weight and Bias in the figure correspond to kernel and bias in the Conv layer
respectively. The title in the figure represents the parameters of the Conv layer to be replaced, and
the abscissa represents the layerd to be replaced(1-25 corresponds to the Conv layer of the network
from front to back). class 1: dark blue,yellow;class 2: orange,baby blue;class 3: gray, green.( The first position of the classes represent the result of parameter replaced, and the second represent the baseline)
}
\label{pspnet-conv}
\end{figure} 

\paragraph{1.2.2 Explore the reason of PSPNet results}~{}

{\bfseries Scale and shift of BN Layer:} 
According to the results shown in Fig3, the network segmentation results are influenced more by RM, RV than RW and RB. So let's assume that $w_j\frac{\mathrm{\Delta}\mu_j}{\sqrt{\sigma_j^2+\varepsilon}} \approx \mathrm{\Delta\ b_j},\ \frac{\sqrt{\sigma_j^2+\varepsilon}}{\sqrt{\sigma_j^2\pm\mathrm{\Delta}\sigma_j^2+\varepsilon}} \approx \alpha_j$.

Plot the figure of $\frac{1}{m}\sum_{j=1}^{m}w_j\frac{\mathrm{\Delta}\mu_j}{\sqrt{\sigma_j^2+\varepsilon}}(\mathrm{\Delta}\mu_j = \sqrt{(\mu_j-\mu_j^\prime)^2}),\frac{1}{m}\sum_{j=1}^{m}\Delta b_j(\Delta b_j=\sqrt{(b_j-b_j^\prime)^2} ),\frac{1}{m}\sum_{j=1}^{m}\frac{\sqrt{\sigma_{j}^{2}+\varepsilon}}{\sqrt{\sigma_{j}^{2} \pm \Delta \sigma_{j}^{2}+\varepsilon}},\frac{1}{m}\sum_{j=1}^{m}\alpha_j  $.

\begin{figure}[H]
\centering
\includegraphics[scale=0.25]{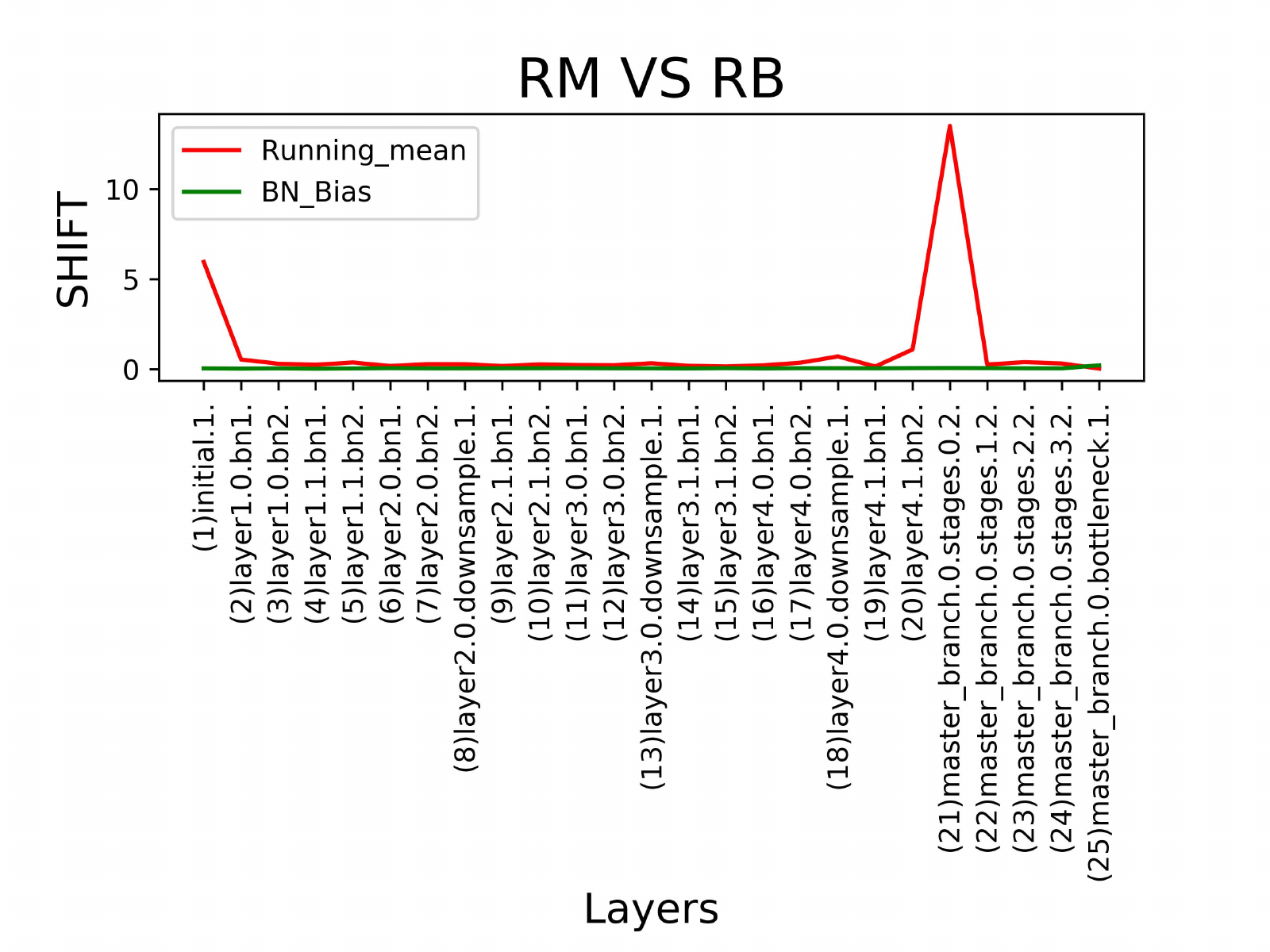}
\includegraphics[scale=0.25]{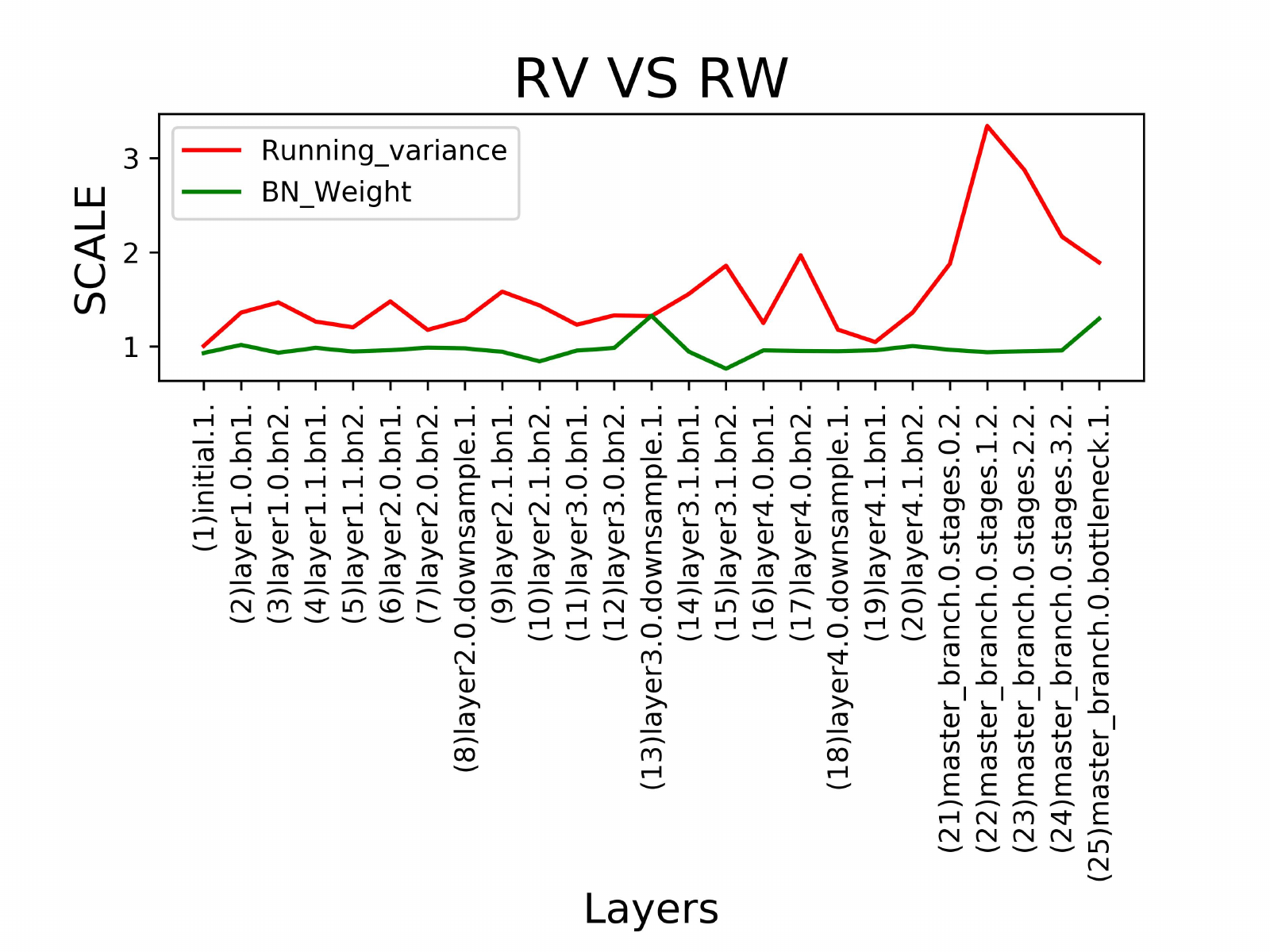}

\caption{PSPNet:Scale and shift in BN layers.Running Mean:$\frac{1}{m}\sum_{j=1}^{m}w_j\frac{\mathrm{\Delta}\mu_j}{\sqrt{\sigma_j^2+\varepsilon}}(\mathrm{\Delta}\mu_j = \sqrt{(\mu_j-\mu_j^\prime)^2})$;BN Bias:$\frac{1}{m}\sum_{j=1}^{m}\Delta b_j(\Delta b_j=\sqrt{(b_j-b_j^\prime)^2} )$;Running variance:$\frac{1}{m}\sum_{j=1}^{m}\frac{\sqrt{\sigma_{j}^{2}+\varepsilon}}{\sqrt{\sigma_{j}^{2} \pm \Delta \sigma_{j}^{2}+\varepsilon}}$;BN Weight:$\frac{1}{m}\sum_{j=1}^{m}\alpha_j $.
}
\label{pspnet-scale-shift}
\end{figure} 
{\bfseries RMSE of Conv layer:} 
\begin{figure}[H]
\centering
\includegraphics[scale=0.25]{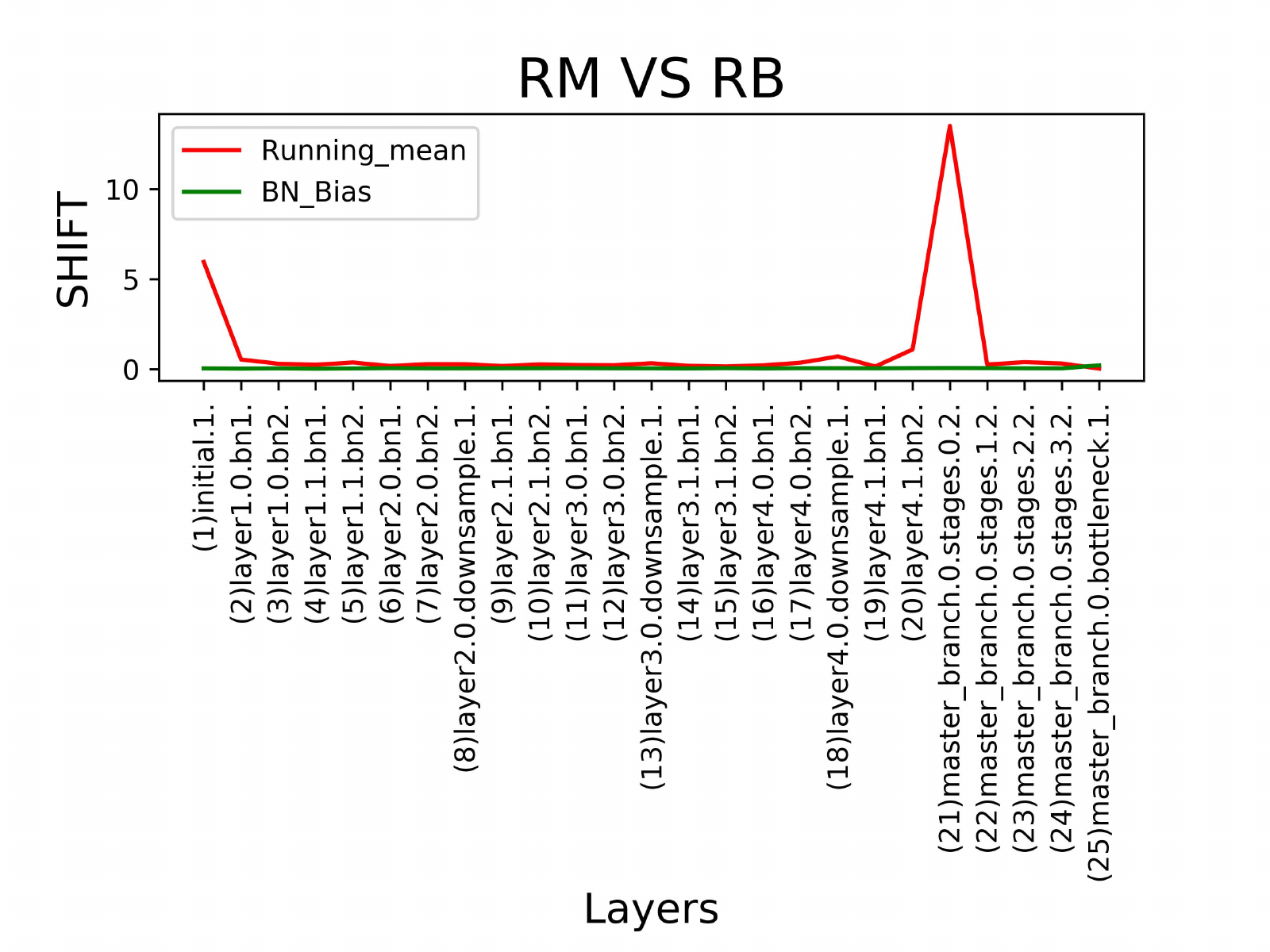}
\includegraphics[scale=0.25]{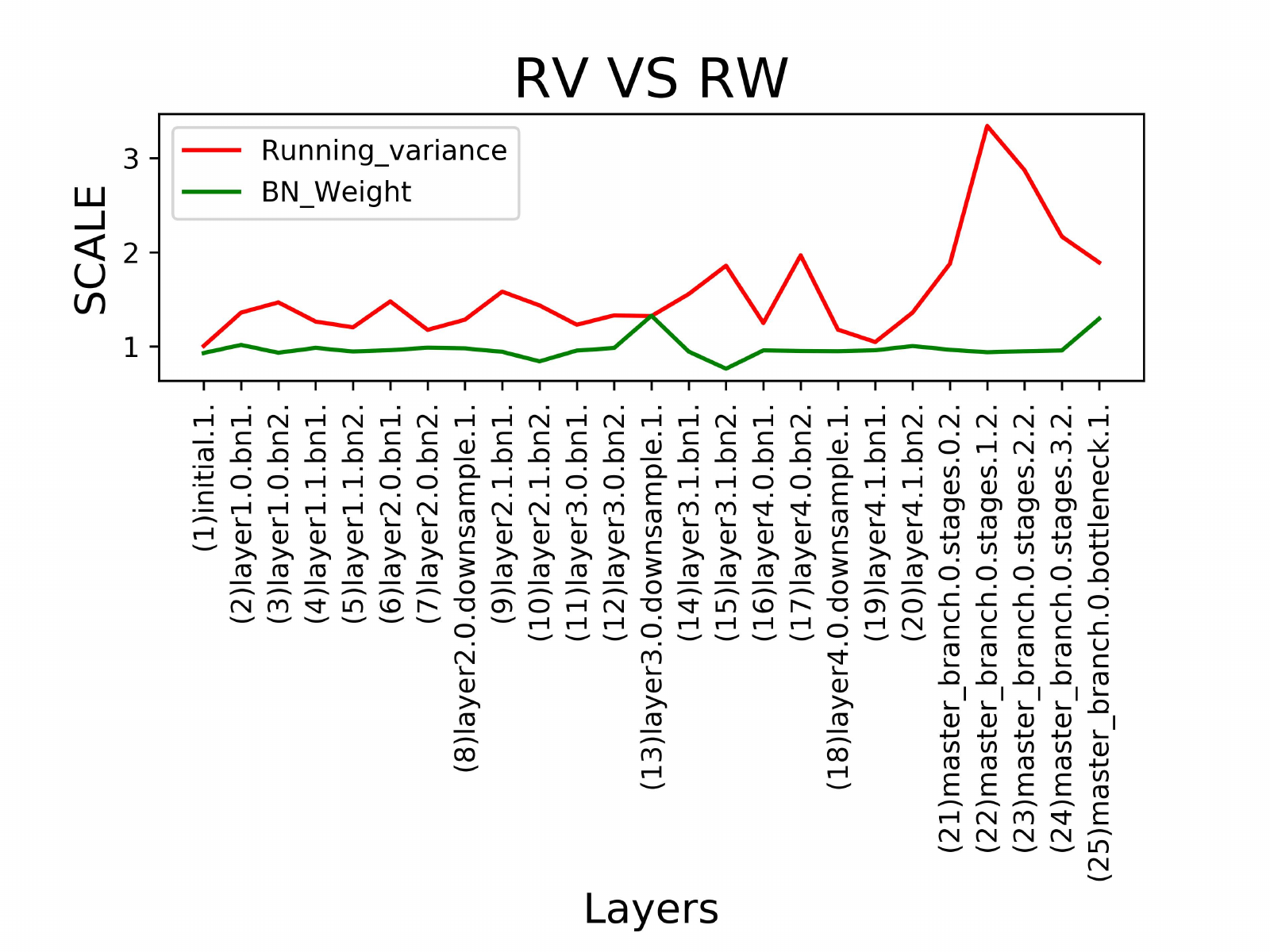}

\caption{PSPNet:The RMSE of W, B, which are the convolutional layers’ parameters, corresponding
to the image segmentation model and image auto-encoding model.
}
\label{pspnet-RMSE}
\end{figure} 

\subsubsection{1.3 Comparative summary between UNet and PSPNet}
According to the comparison between UNet and PSPNet parameter replacement results, it is found that the influence of parameter reuse on UNet is greater than that of PSPNet. In the BN layer, RM and RV have greater influence on the result than RW and RB. Moreover, from the scale and shift results in the images, it conforms to the reasoning in the Theory.

\subsection{2.Different tasks and different data sets differ in network parameters}

Network parameter reuse, the intuitive requirement is that the parameters in the network are generic. This paper believes that it is also related to the difference of parameters. In order to explore the differences between network parameters of different models, this section calculates the RMSE between the segmentation and auto-encoder corresponding layer parameters of T1,RVSC and ACDC data sets.

\begin{figure}[H]
\centering
\includegraphics[scale=0.125]{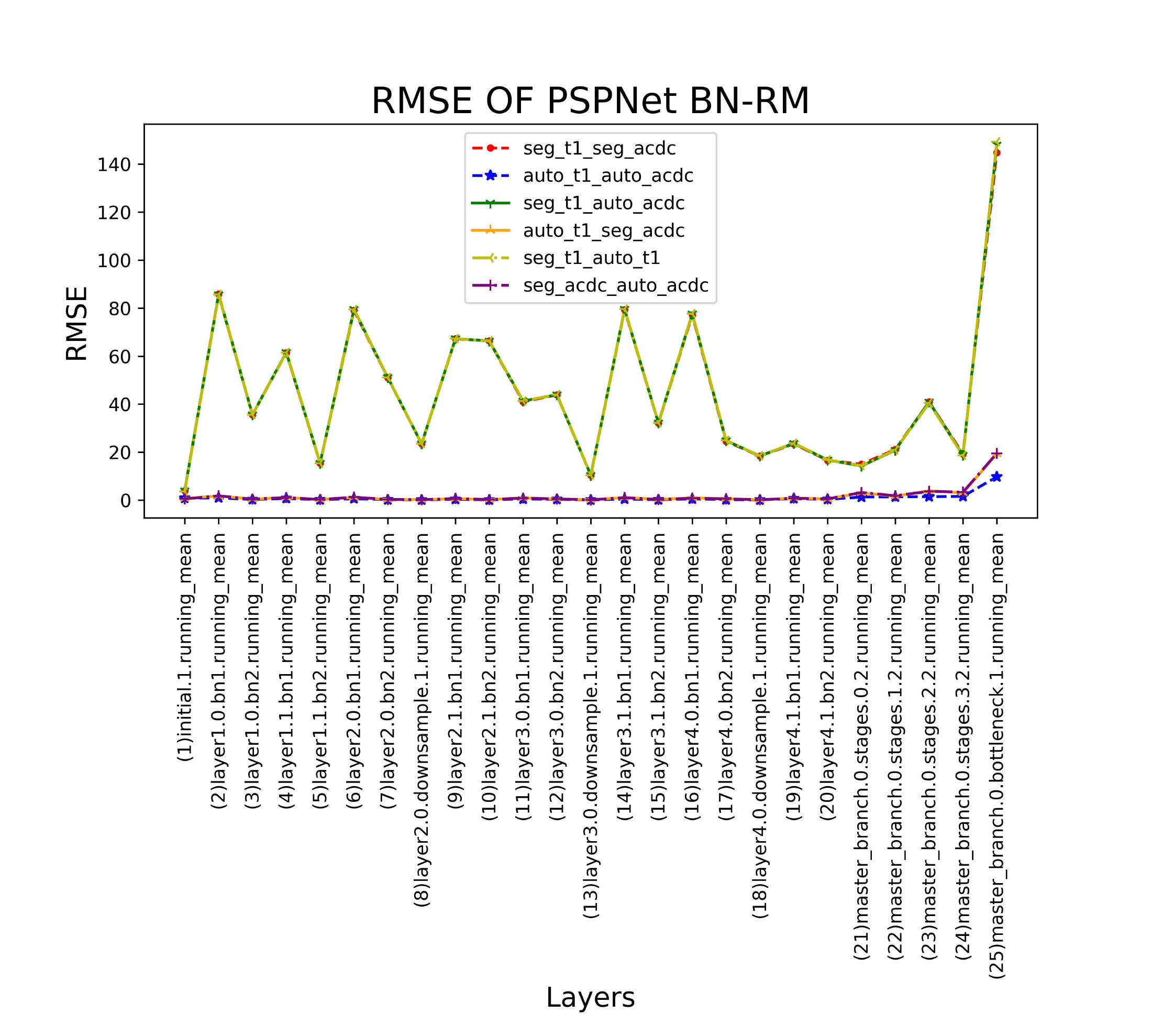}
\includegraphics[scale=0.125]{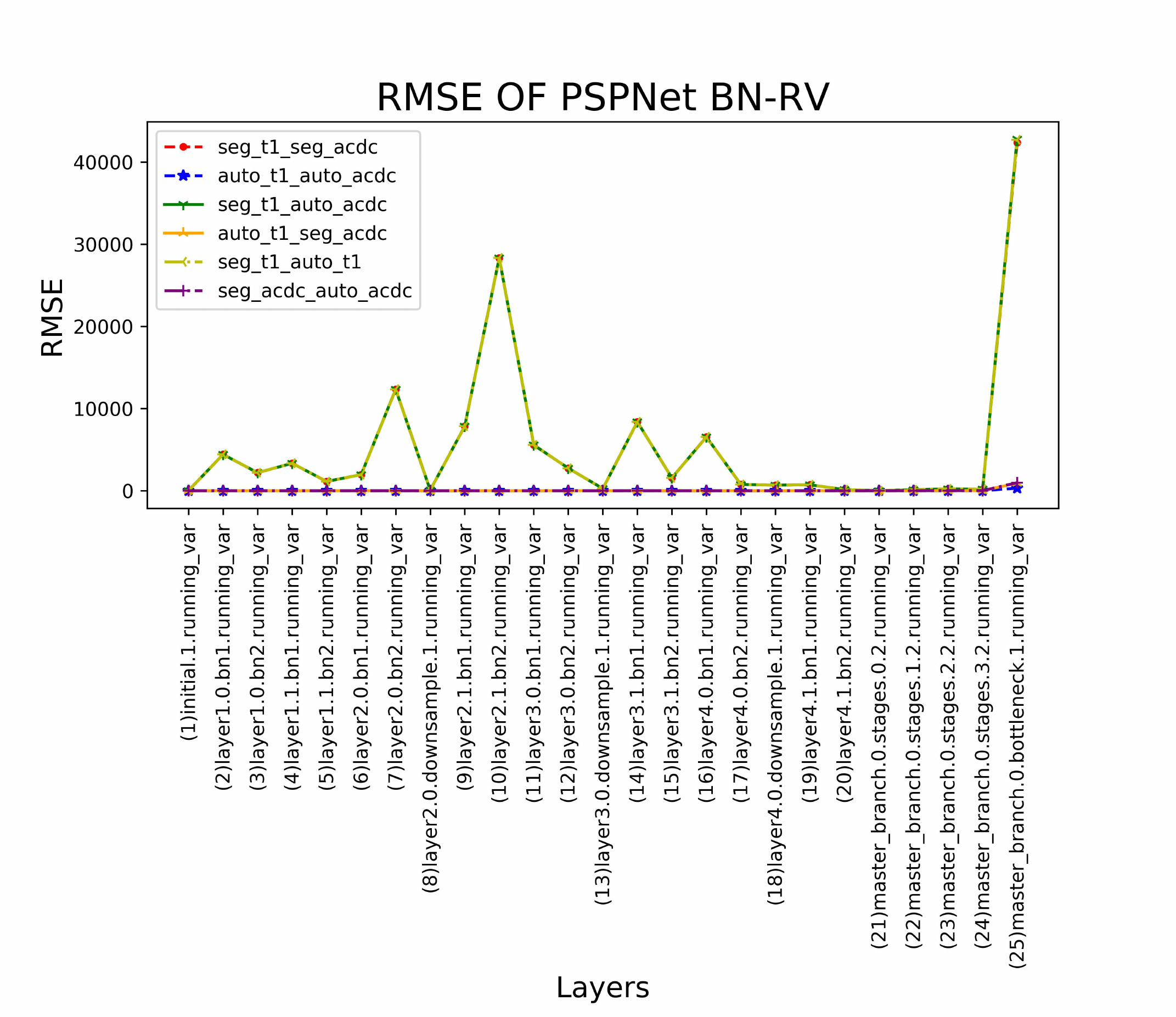}

\includegraphics[scale=0.138]{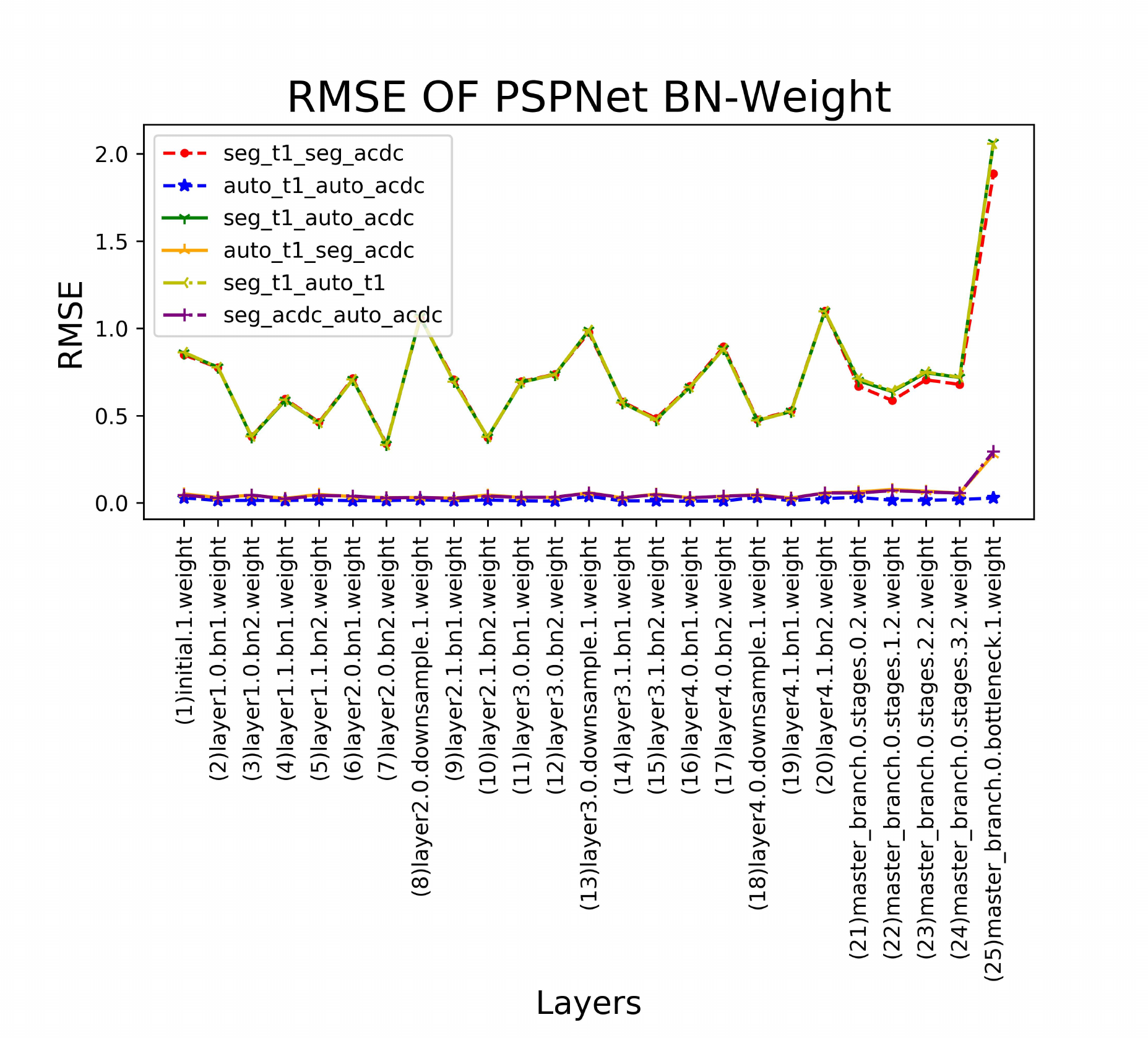}
\includegraphics[scale=0.14]{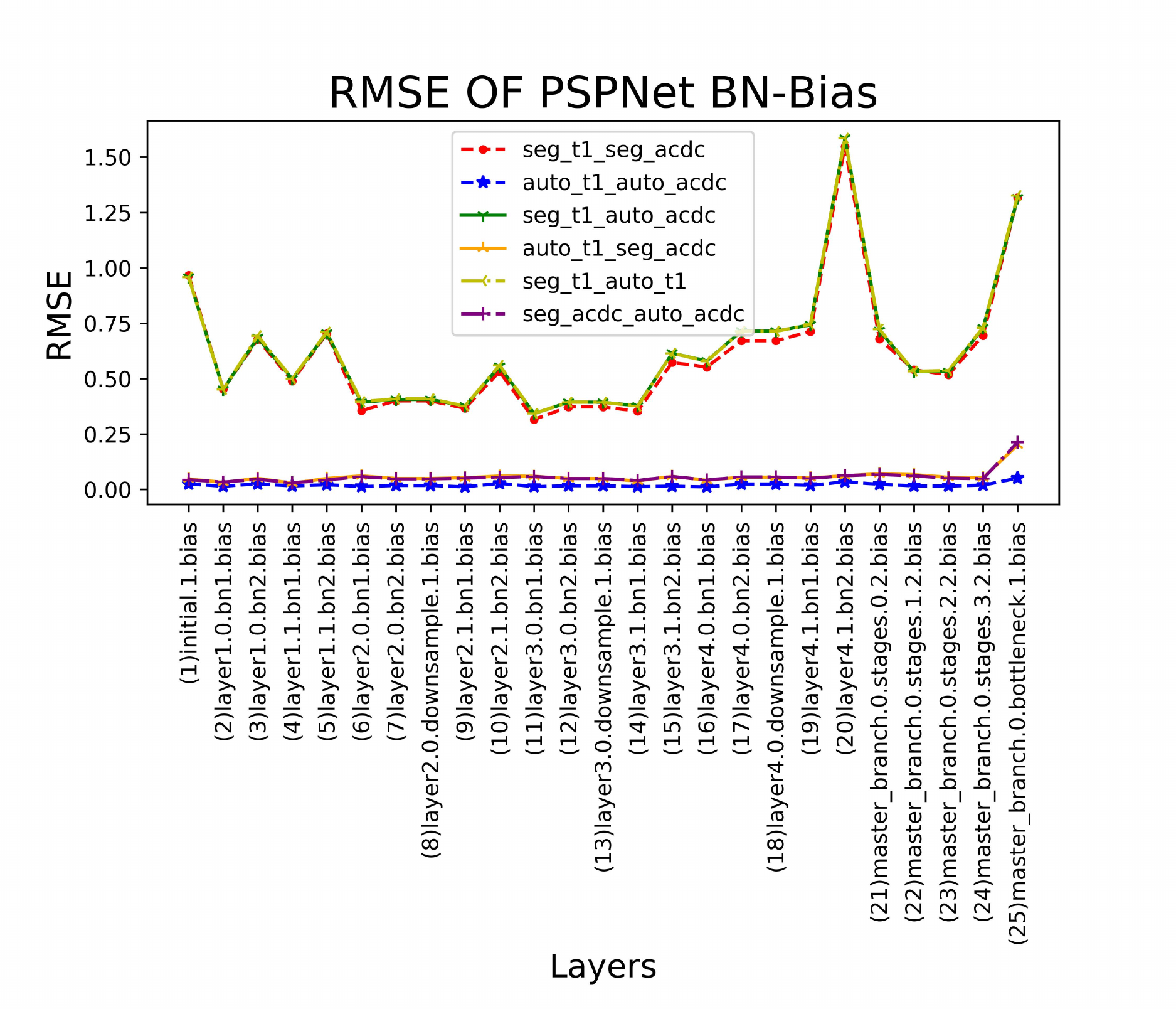}

\includegraphics[scale=0.15]{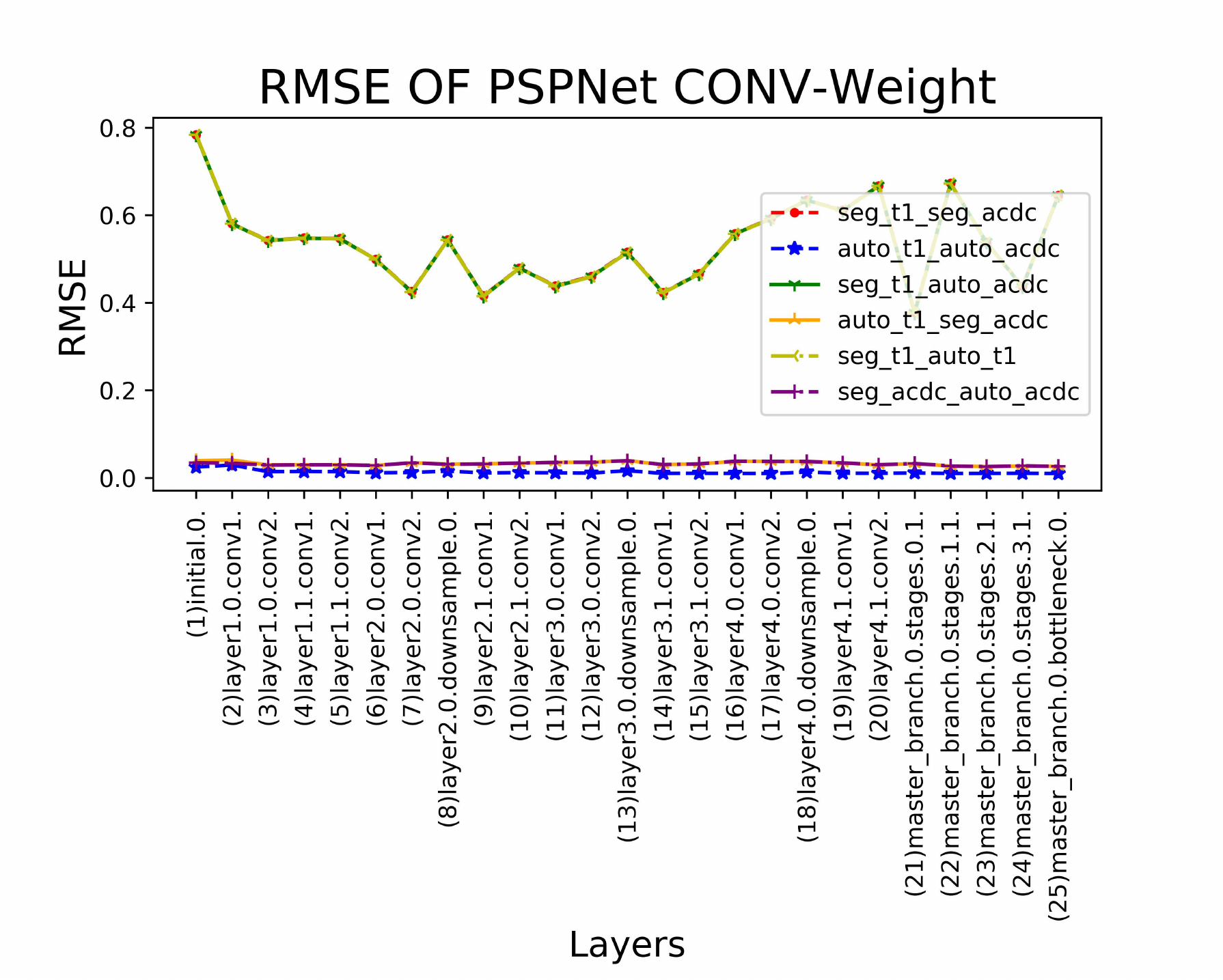}
\includegraphics[scale=0.15]{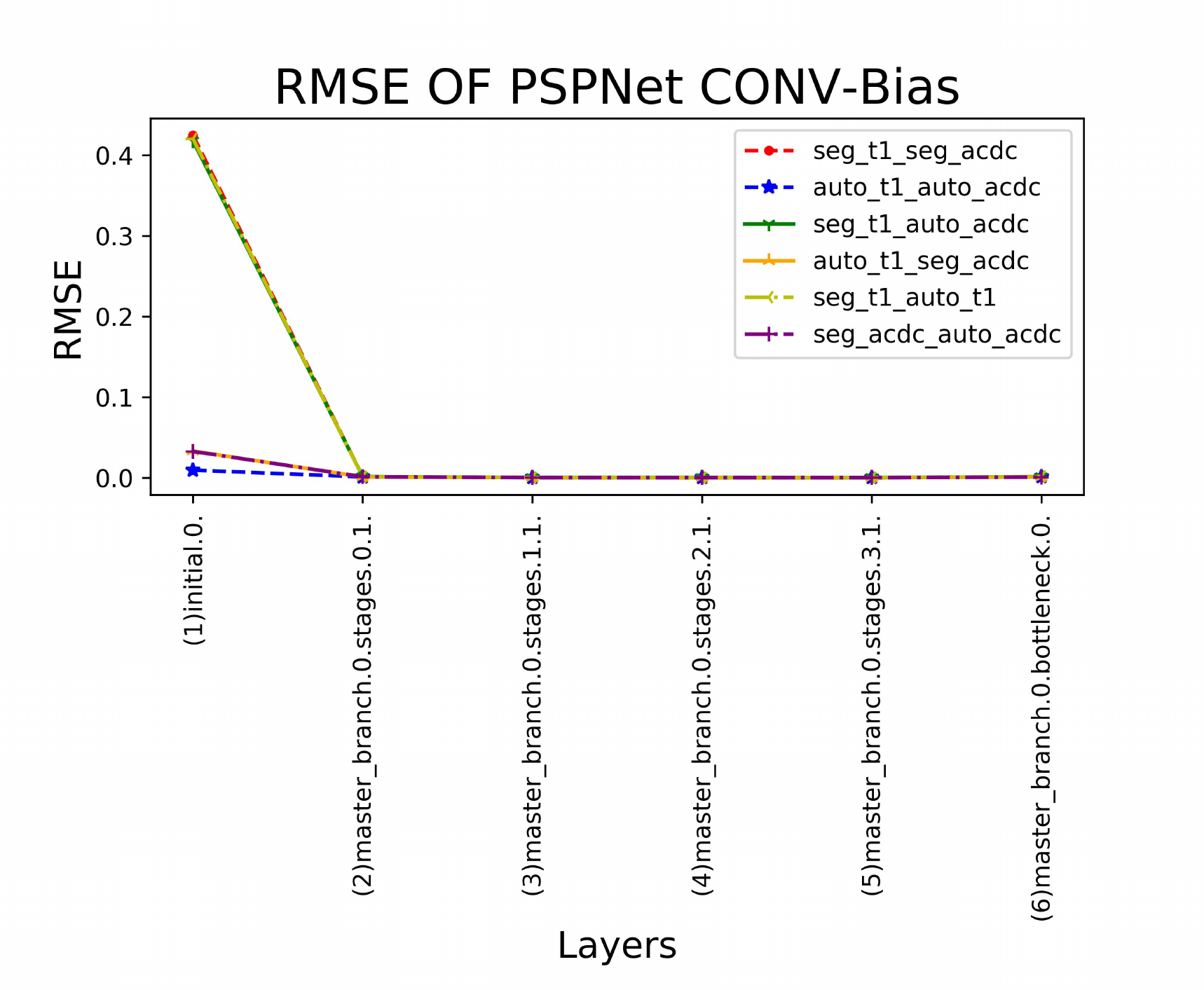}

\caption{Different models of PSPNet(RESnet-18)network RMSE.RMSE is calculated by two sets of models, seg\_T1,auto\_T1, seg\_ACDC and auto\_ACDC in the annotation respectively represents the auto-encoder and segmentation model of T1 data set and the auto-encoder and segmentation model of ACDC data set.
}

\label{pspnet-RMSE_T1_ACDC}
\end{figure} 

\begin{figure}[H]
\centering
\includegraphics[scale=0.125]{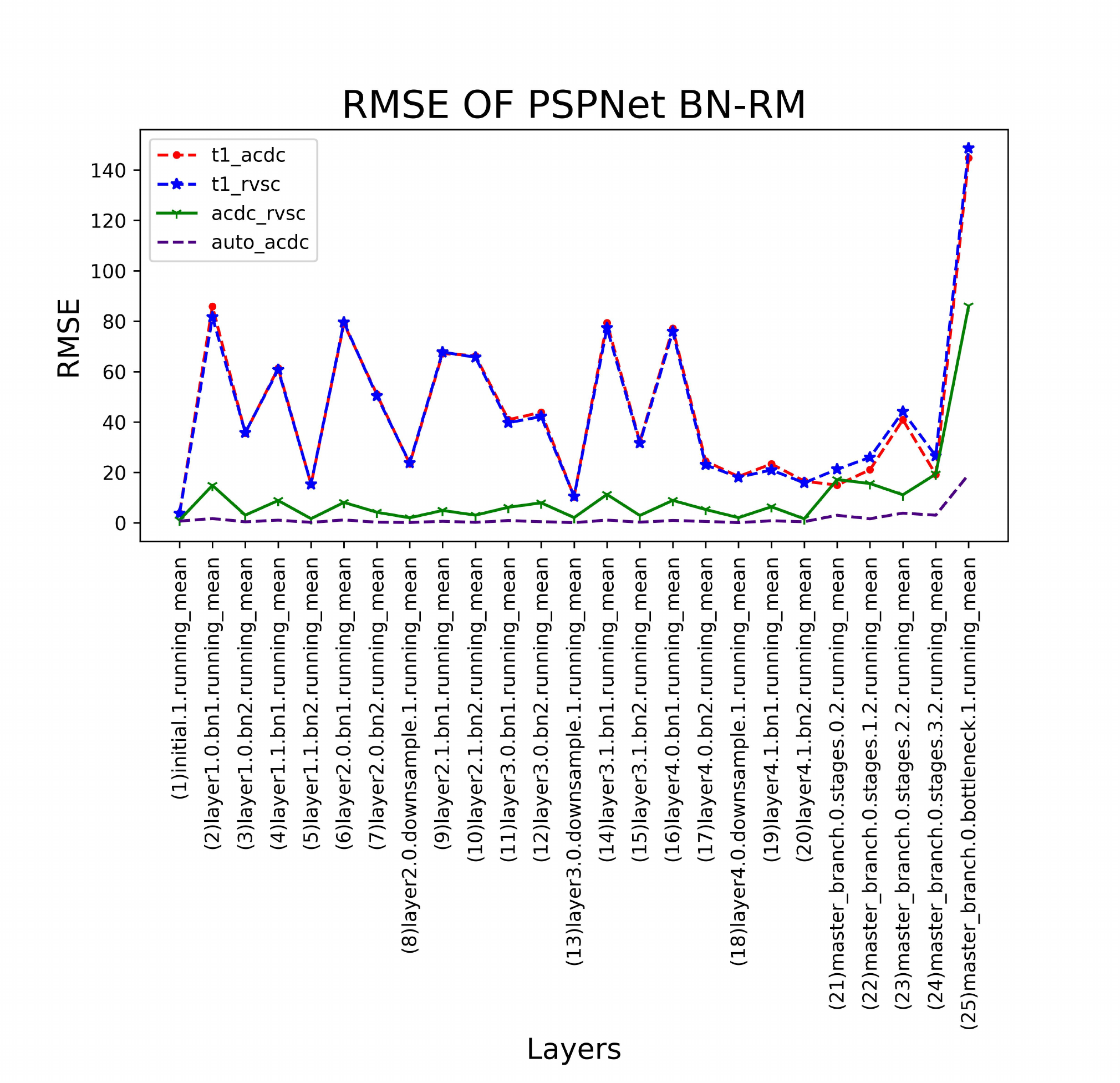}
\includegraphics[scale=0.125]{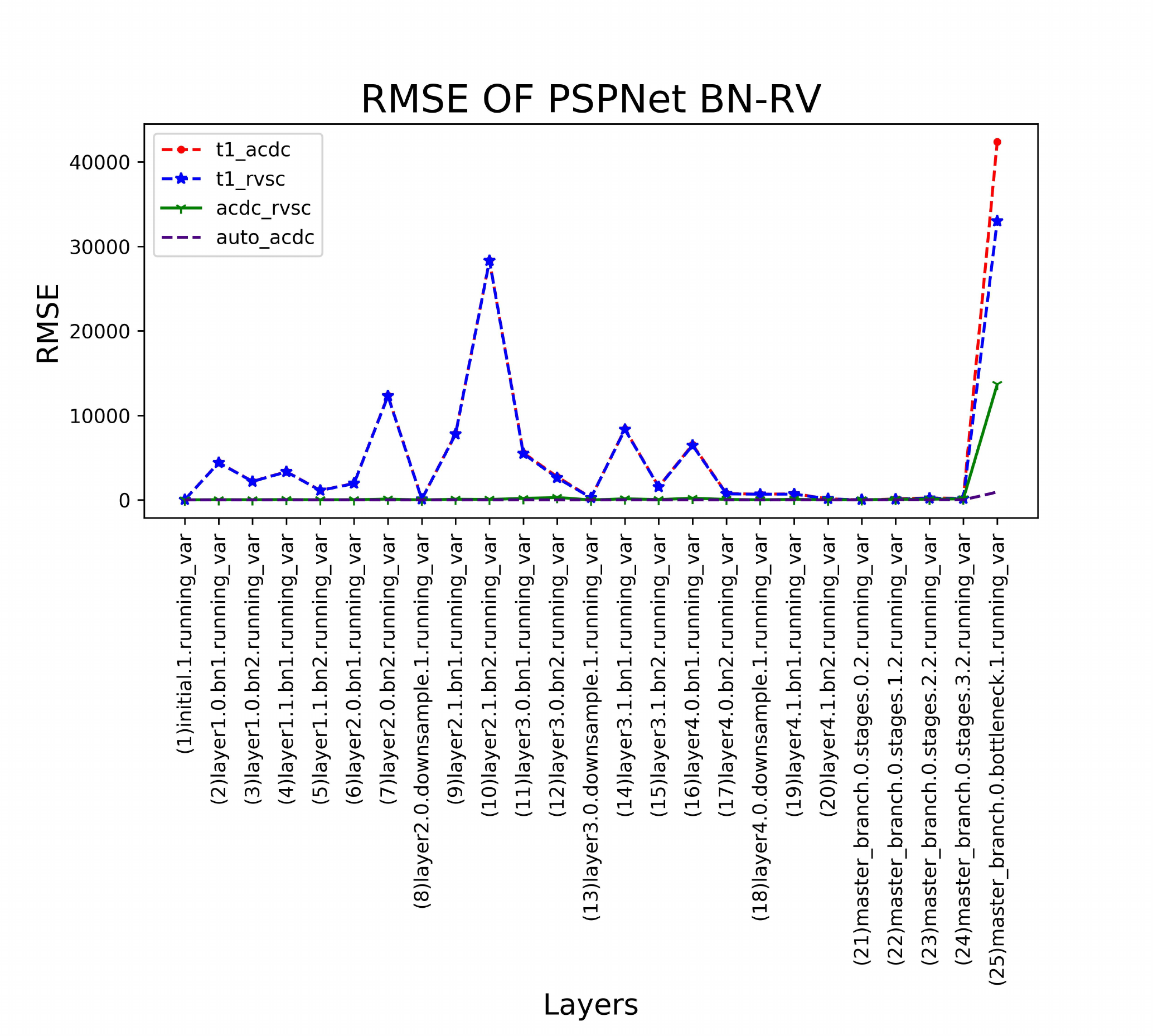}

\includegraphics[scale=0.138]{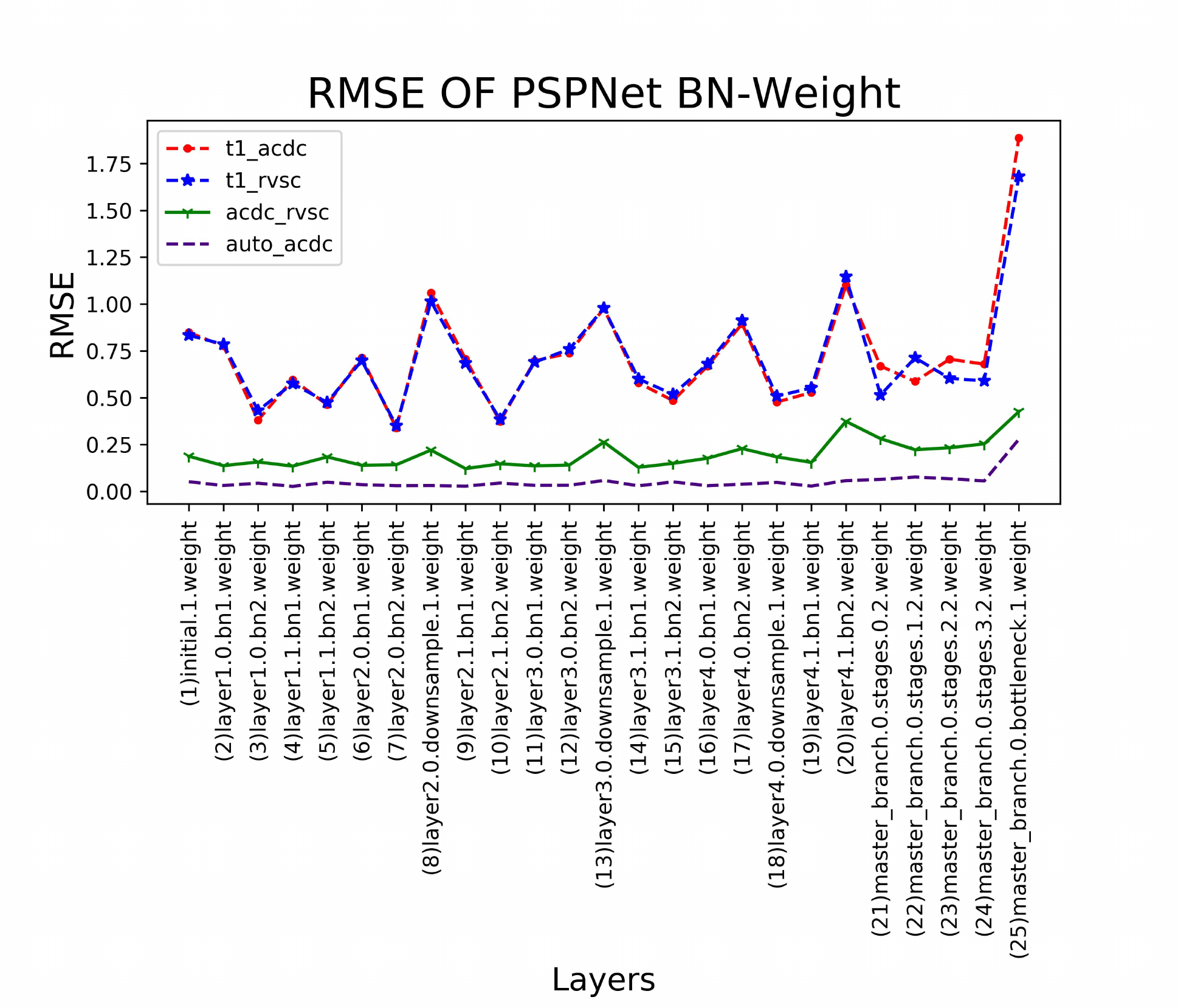}
\includegraphics[scale=0.14]{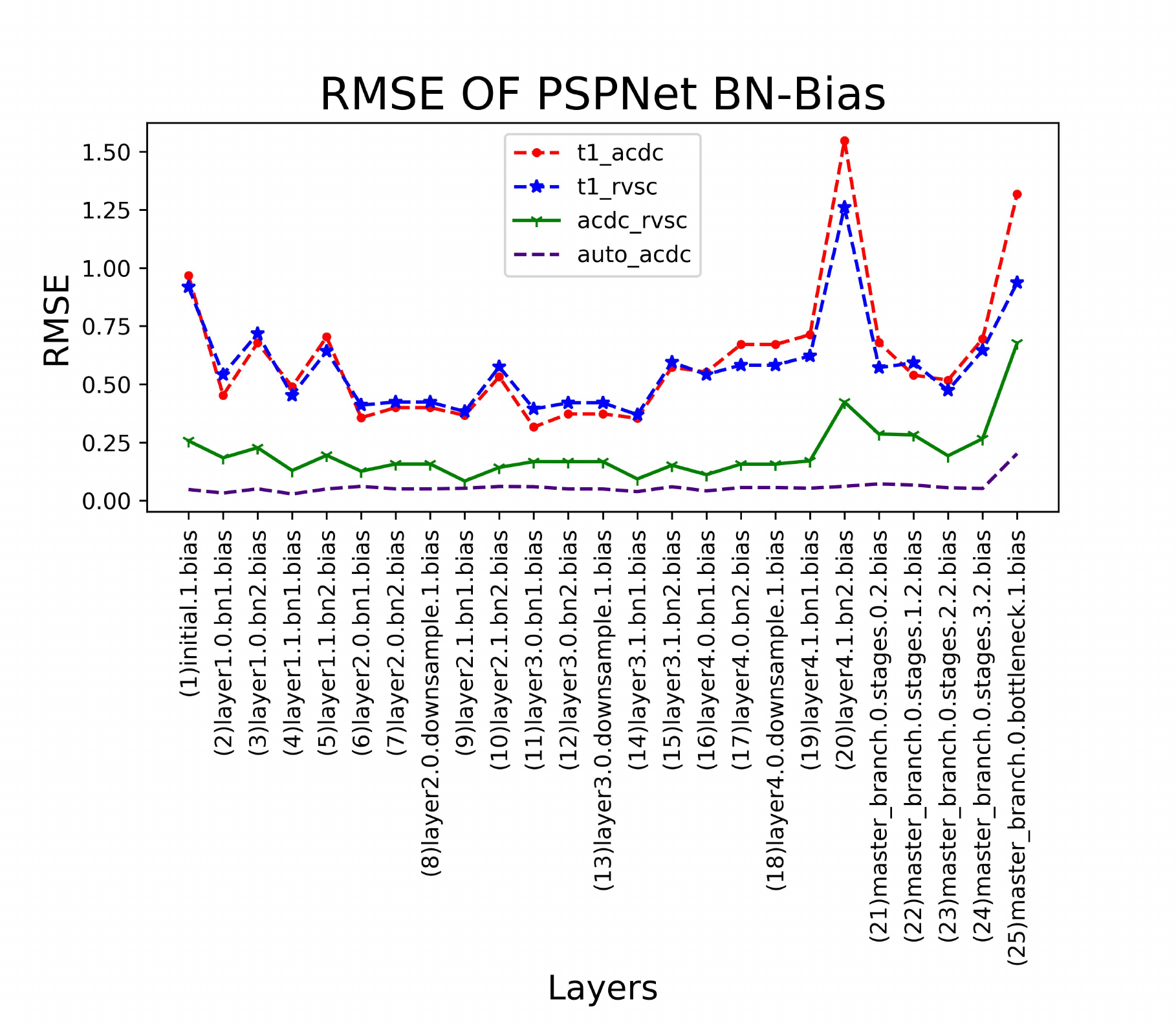}

\includegraphics[scale=0.15]{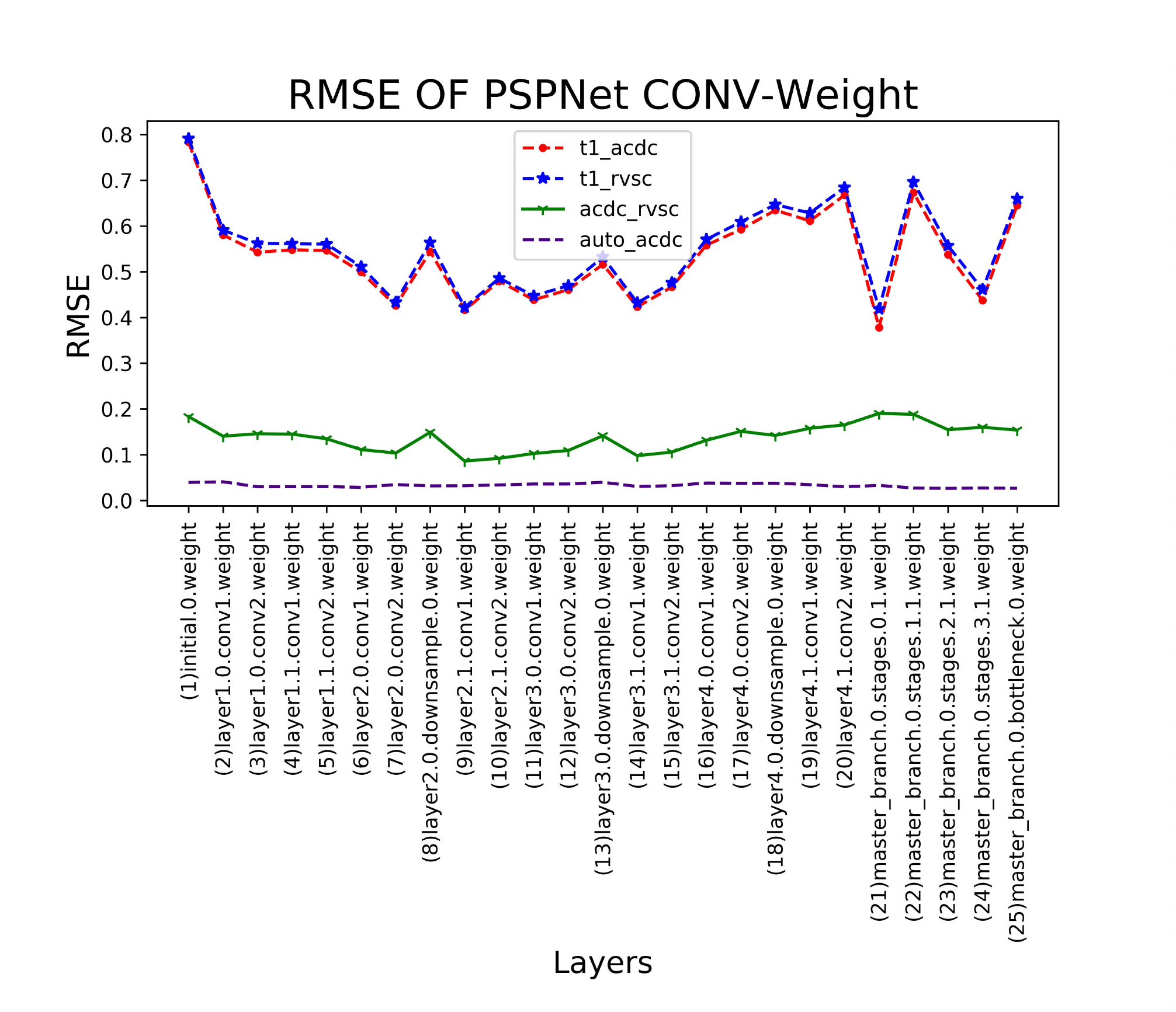}
\includegraphics[scale=0.15]{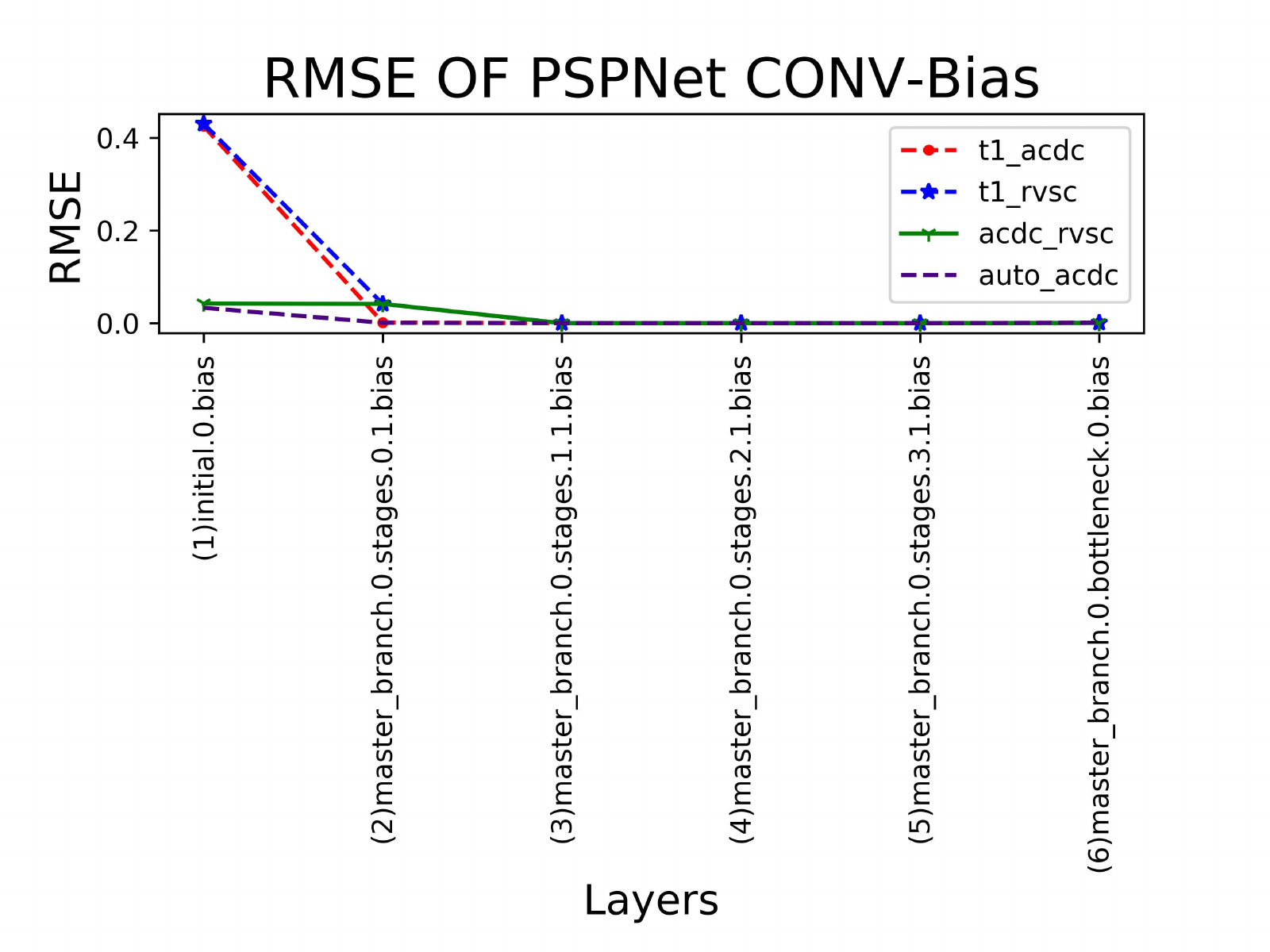}

\caption{RMSE is calculated by four models of PSPNet (RESnet-18) network. T1,RVSC and ACDC in the annotation respectively represent the image segmentation model of T1,RVSC and ACDC data sets.auto represents the image auto-encoder model of T1 dataset.
}

\label{pspnet-RMSE_T1_ACDC_RVSC}
\end{figure} 
RMSE images in Fig11 were roughly divided into two groups, all the RMSE values with large differences were related to the T1 data set segmentation model, while the others have small differences. Fig16 images are roughly divided into three groups. RMSE values of corresponding layer parameters of T1 image auto-encoder model and ACDC image segmentation model are the minimize differences.  

The above phenomena can be summarized as follows:
A. Same data set, different tasks (segmentation and image auto-encoder), and the RMSE is uncertain.
B. Different data sets, different tasks, and the RMSE is uncertain.
C. Different data sets, same task, and the RMSE is uncertain.
According to the above results, there is little difference between T1 data set auto-encoder parameters and ACDC data set image segmentation parameters, so it can be inferred that T1 data set image auto-encoder parameters can be reused in ACDC segmentation task.

\subsubsection{3 Parametric reuse experiment}~{}   

In order to verify the conclusion of the above network parameter reuse experiment, this section reuses the parameters learned from T1 image auto-encoder, image segmentation and RVSC image segmentation into ACDC image segmentation. The experiment is divided into three parts. The first part was randomly initialized without loading any parameters. The second part loaded the T1 auto-encoder, image segmentation and RVSC image segmentation parameters and then fixed the inferred reusable parameters to train. In the third part, T1 auto-encoder, image segmentation and RVSC image segmentation parameters were loaded, and fine-tuned the inferred reusable parameters.  

\paragraph{3.1 Result of PSPNet}~{}  

According to the RMSE of T1 auto-encoder and ACDC segmentation, it is speculated that parameters non-reusable layers of PSPNet (ResNet18) is RM: 1,17;W: 1,6,11,12,16,17,19,20;B: 1. All other parameters were reused. If each variable is treated independently and each layer is treated independently, a total of 92\% of the parameters were reused in this experiment. Other models do the same operation.

\begin{table}[H]
    \centering
    \caption{Segmentation results of PSPNet (RESnet-18) network reuse T1 data set auto-encoder and segmentation parameters on the ACDC data set.}
    \scalebox{0.5}{
    \begin{tabular}{|l|l|l|l|l|}
        models & class-0 & class-1 & class-2 & class-3 \\ 
        10-samples don't load pretrained parameters & 0.99 & 0.62 & 0.60 & 0.70 \\ 
        10-sample T12ACDC seg2seg freeze & 0.99 & 0.47 & 0.56 & 0.67 \\ 
        10-sample T12ACDC seg2seg Do not freeze & 1.00 & 0.48 & 0.55 & 0.71 \\ 
        10-sample T12ACDC auto2seg freeze & 0.99 & 0.70 & 0.70 & 0.84 \\ 
        10-sample T12ACDC auto2seg Do not freeze & 1.00 & 0.75 & 0.75 & 0.88 \\ 
        10-sample RVSC2ACDC freeze & 0.99 & 0.57 & 0.53 & 0.72 \\ 
        10-sample RVSC2ACDC Do not freeze & 0.99 & 0.54 & 0.55 & 0.74 \\ 
        50-samples don't load pretrained parameters & 1.00 & 0.81 & 0.82 & 0.90 \\ 
        50-sample T12ACDC seg2seg freeze & 0.99 & 0.69 & 0.75 & 0.84 \\ 
        50-sample T12ACDC seg2seg Do not freeze & 0.99 & 0.67 & 0.75 & 0.84 \\ 
        50-sample T12ACDC auto2seg freeze & 1.00 & 0.84 & 0.83 & 0.90 \\ 
        50-sample T12ACDC auto2seg Do not freeze & 1.00 & 0.86 & 0.86 & 0.93 \\ 
        50-sample RVSC2ACDC freeze & 0.99 & 0.70 & 0.70 & 0.86 \\ 
        50-sample RVSC2ACDC Do not freeze & 0.99 & 0.67 & 0.74 & 0.82 \\ 
    \end{tabular}
    }
\end{table}

In the table, 10 and 50 represent the number of samples (in order to ensure the uniformity of the experiment, the validation set is the same people as the first part, and 10 people in the training set are randomly selected from 50 people in the original training set).Don't load pretrained parameters means random initialization. Freeze means freeze the layers that loads the pre-trained parameters and Do not Freeze means the pre-trained parameters are trained with the network training. seg2seg means transfer segmentation parameter to ACDC segmentation, auto2seg means transfer auto-coder parameter to ACDC segmentation. T12ACDC indicates that the pre-trained parameters are derived from the T1 data set. RVSC2ACDC representation and pre-trained parameters are derived from RVSC. The above results also explain why the results sometimes vary greatly when papers are reproduced.  

According to section2, we draw the conclusion that:  

a. The transfer learning results are not always good. It is required that the parameters of the original ideal model to be transferred should be similar enough to the parameters of the ideal model in this data set.  

b. The essence of transfer learning lies in the similarity of transfer parameters to the ideal parameters, which is not derived from the similarity of data set or task, and needs to be analyzed on a case-by-case basis.

\paragraph{3.2 RMSE of parameters}~{} 

Calculate RMSE value of corresponding layers of ACDC segmentation models after parameters reuse.
RMSE values of parameters after T1 image auto-encoder model parameters are reused by ACDC image segmentation models:

\begin{figure}[H]
\centering
\includegraphics[scale=0.125]{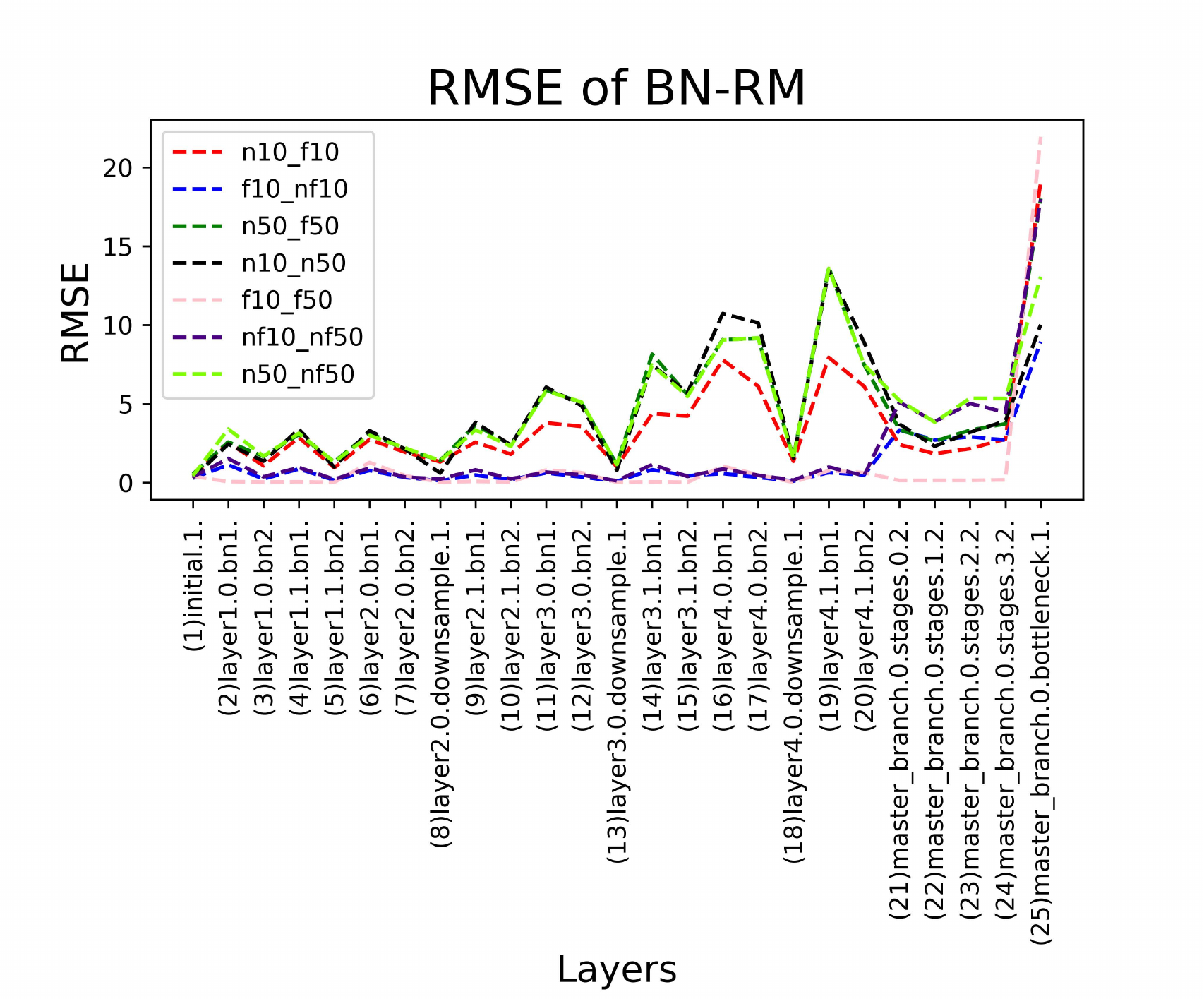}
\includegraphics[scale=0.125]{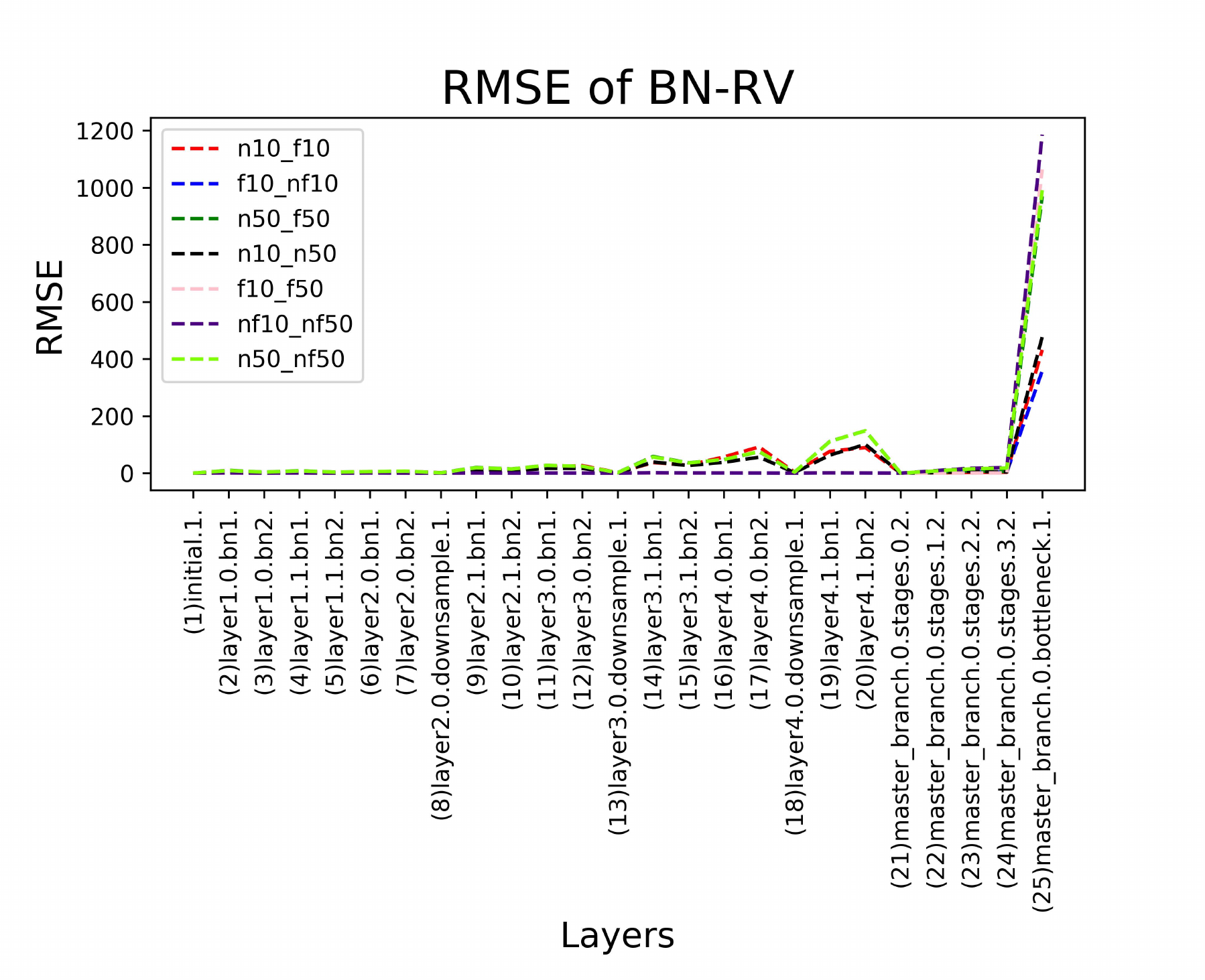}

\includegraphics[scale=0.138]{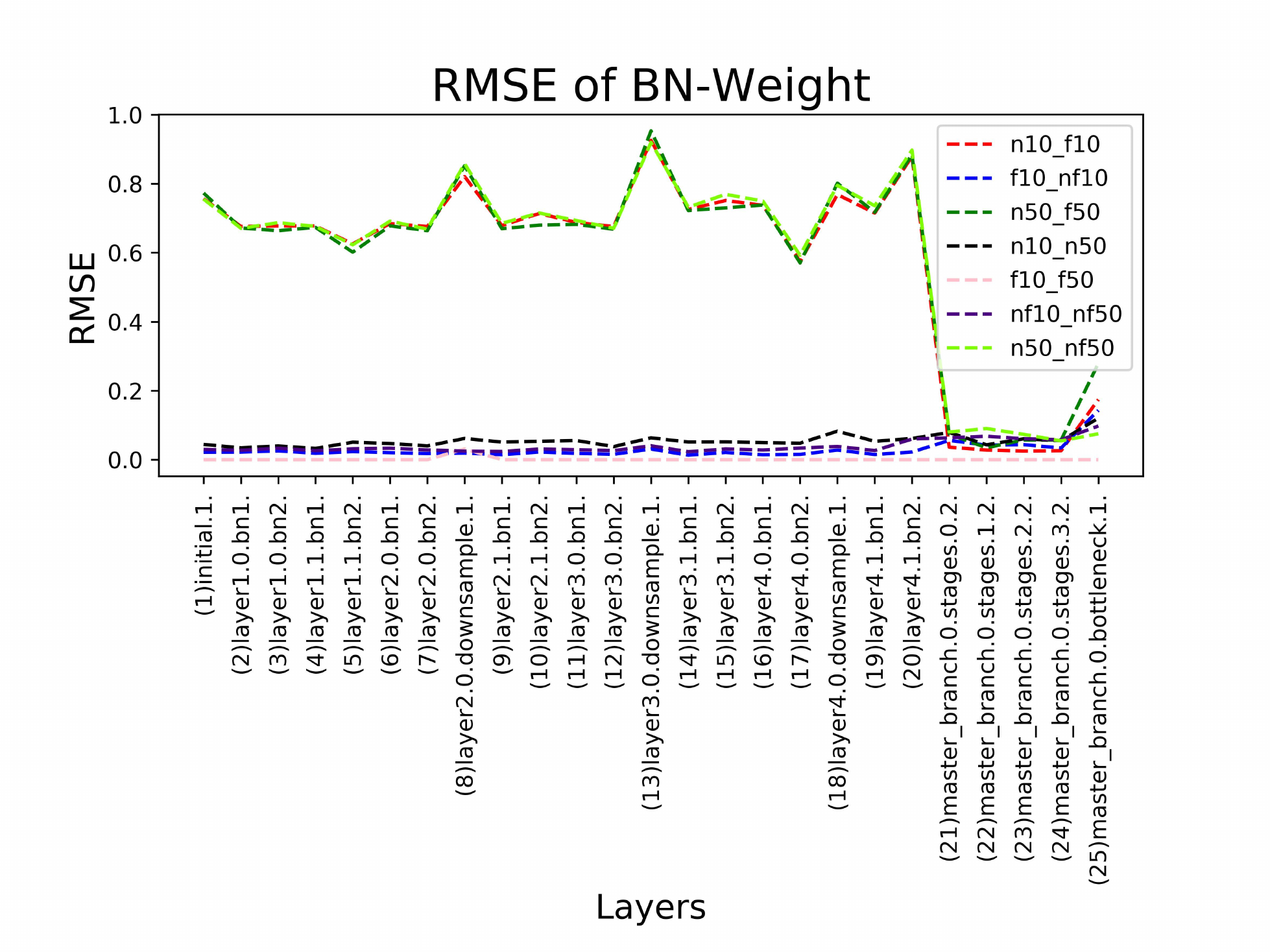}
\includegraphics[scale=0.14]{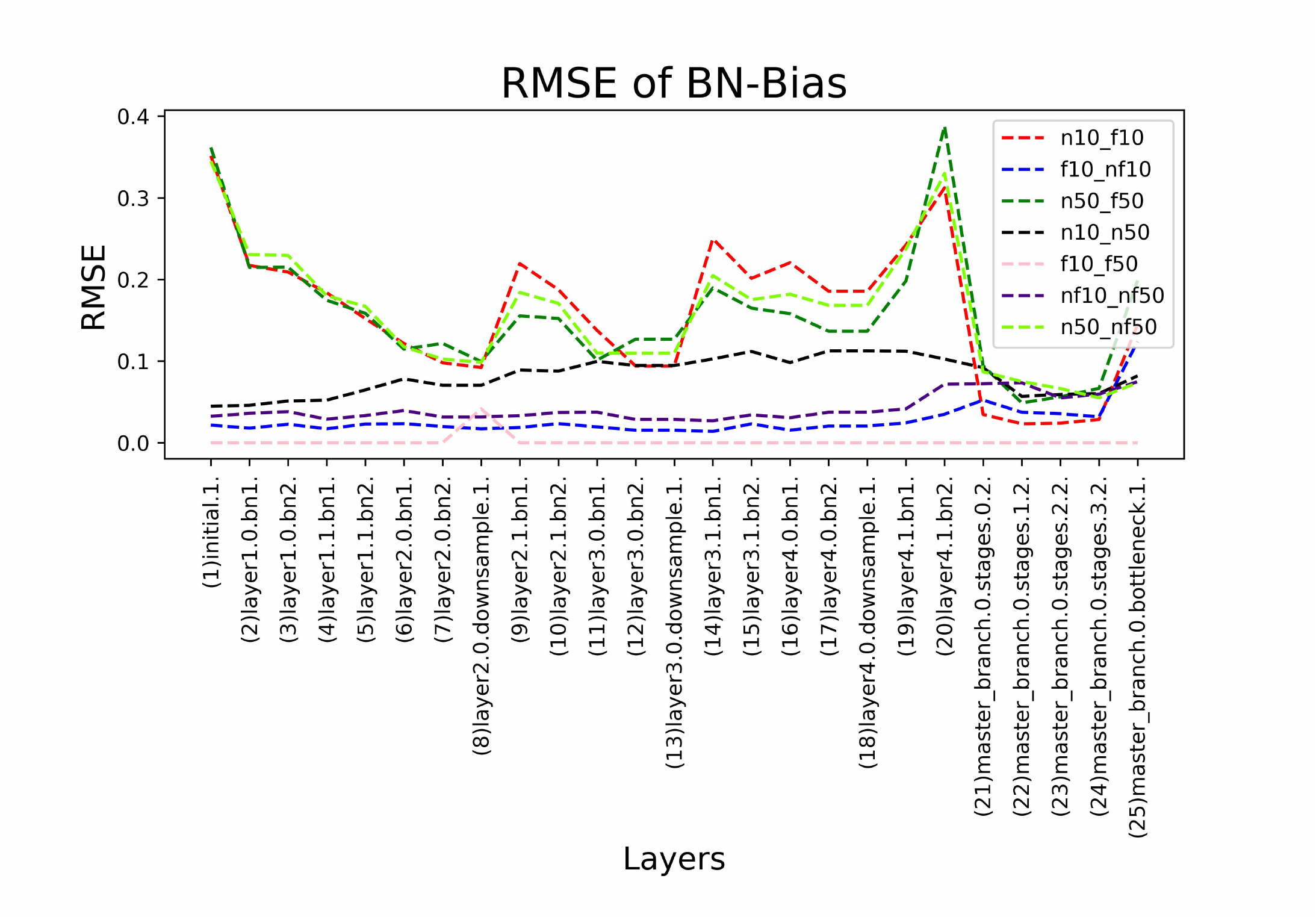}

\includegraphics[scale=0.15]{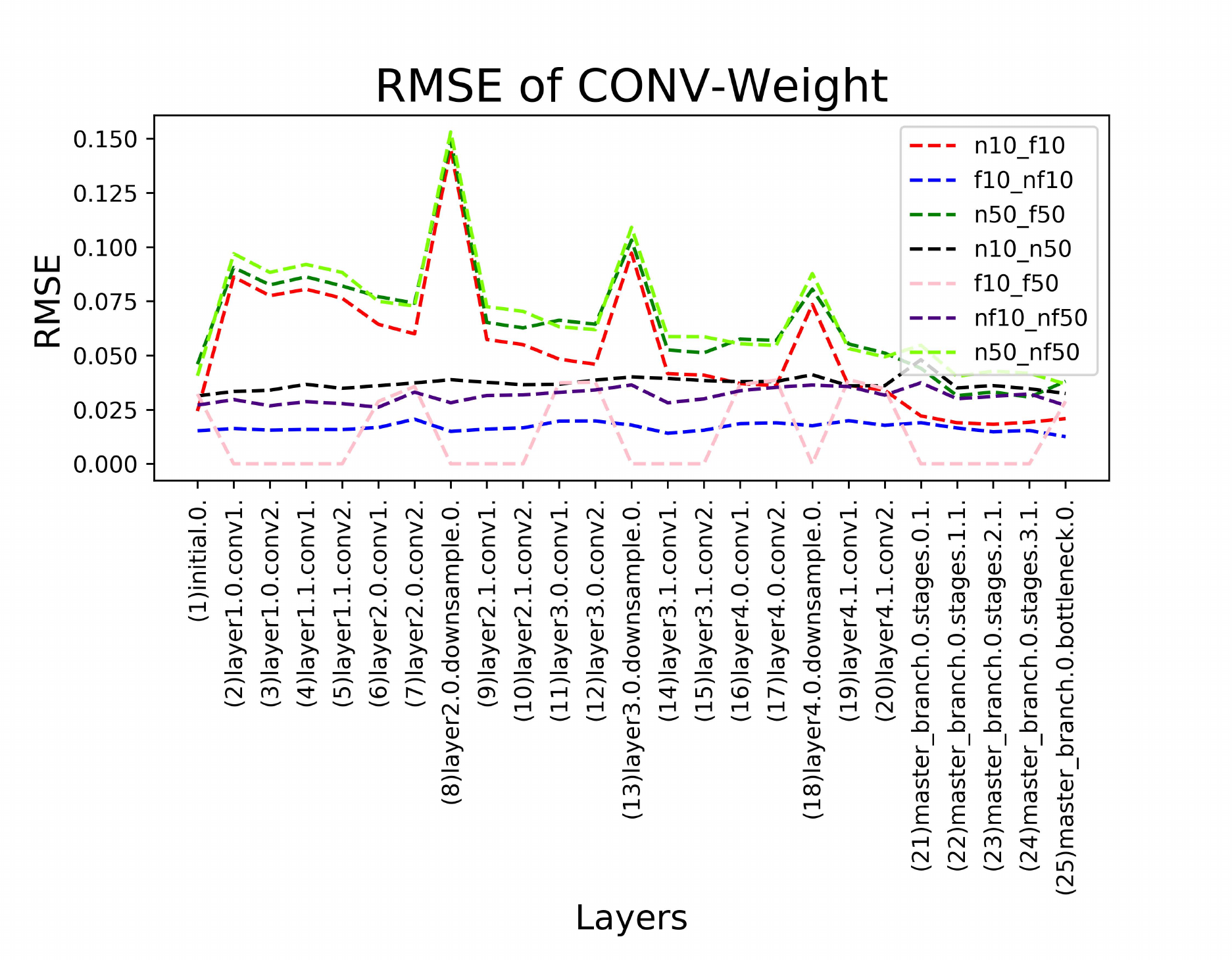}
\includegraphics[scale=0.15]{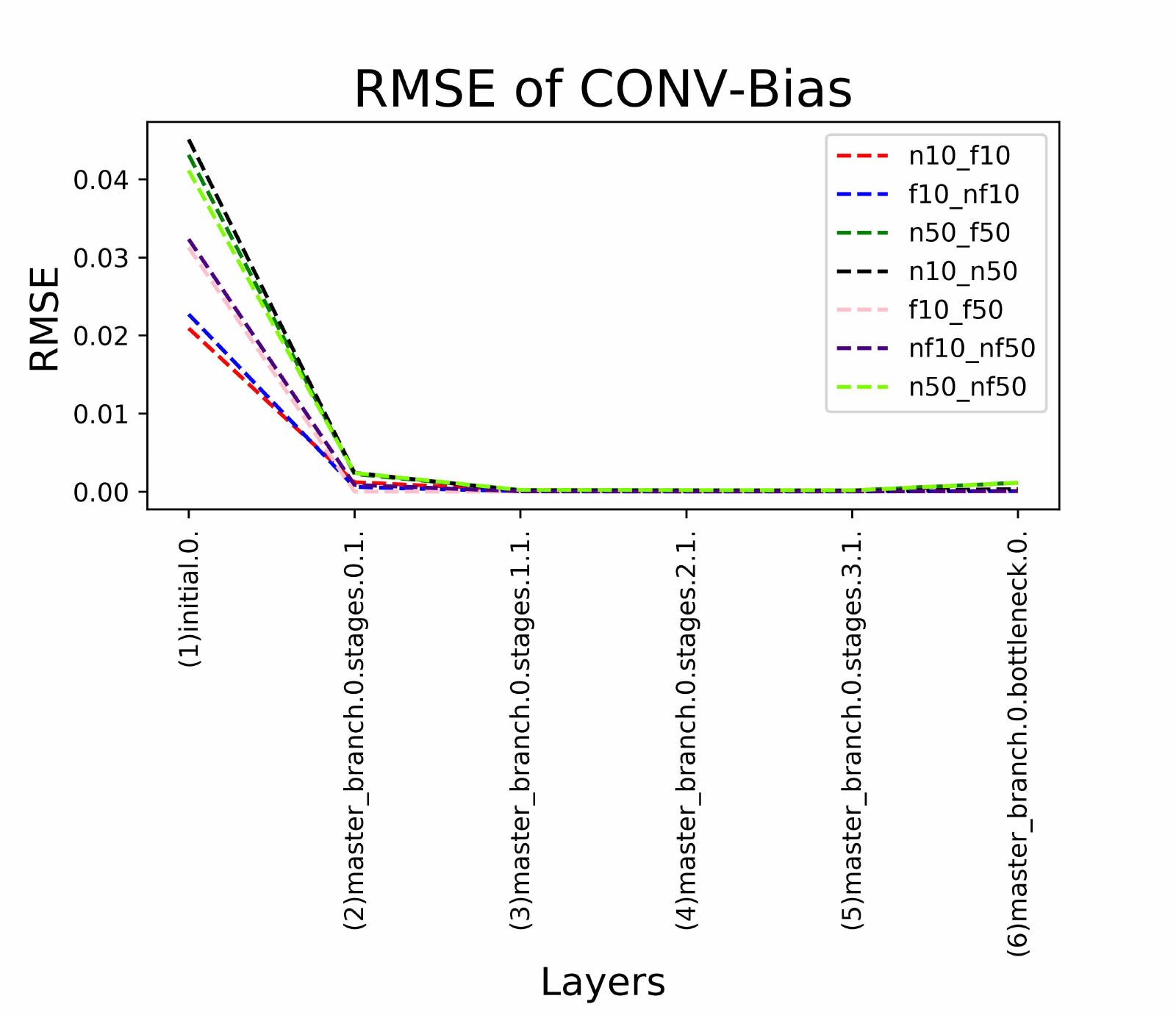}

\caption{PSPNet: After T1 image auto-encoder model parameters are reused by ACDC image segmentation models, the difference of parameters of corresponding layer of each model. The horizontal axis represents the layers of the model, the vertical axis represents the RMSE values, and the title represents the variable whose RMSE values are computed. The icons represent different models. n: Do not load pretrained parameters; f: Freeze loaded pre-trained parameters; nf: Do not freeze loaded pre-trained parameters.10,50: The number of patients is used for training.
}

\label{pspnet-RMSE_T1_ACDC_RVSC}
\end{figure} 

RMSE of T1 nf50 model (The model that ACDC image segmentation reused T1 image auto-encoder parameters) and ACDC reused RVSC image segmentation models parameters:

\begin{figure}[H]
\centering
\includegraphics[scale=0.125]{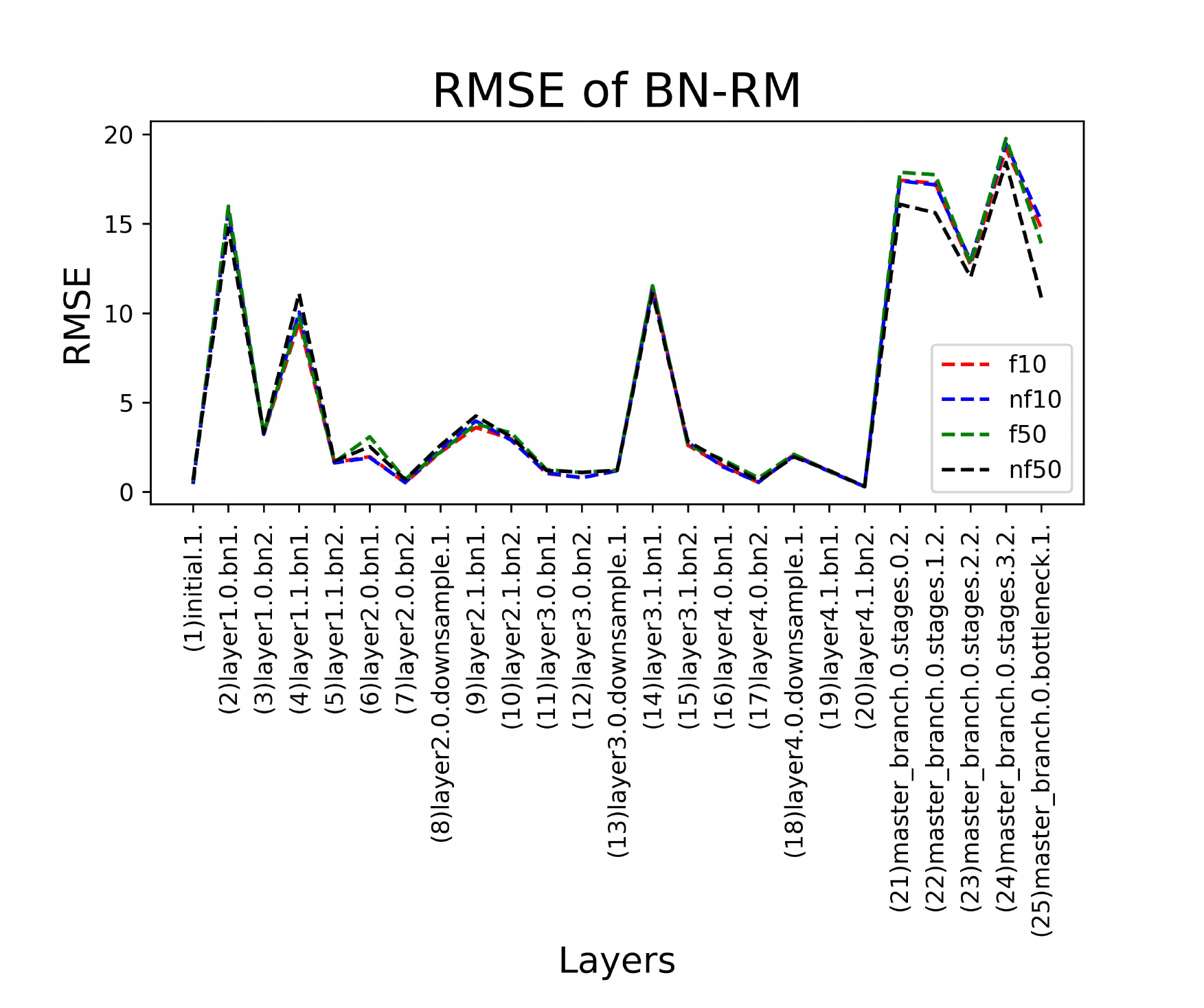}
\includegraphics[scale=0.125]{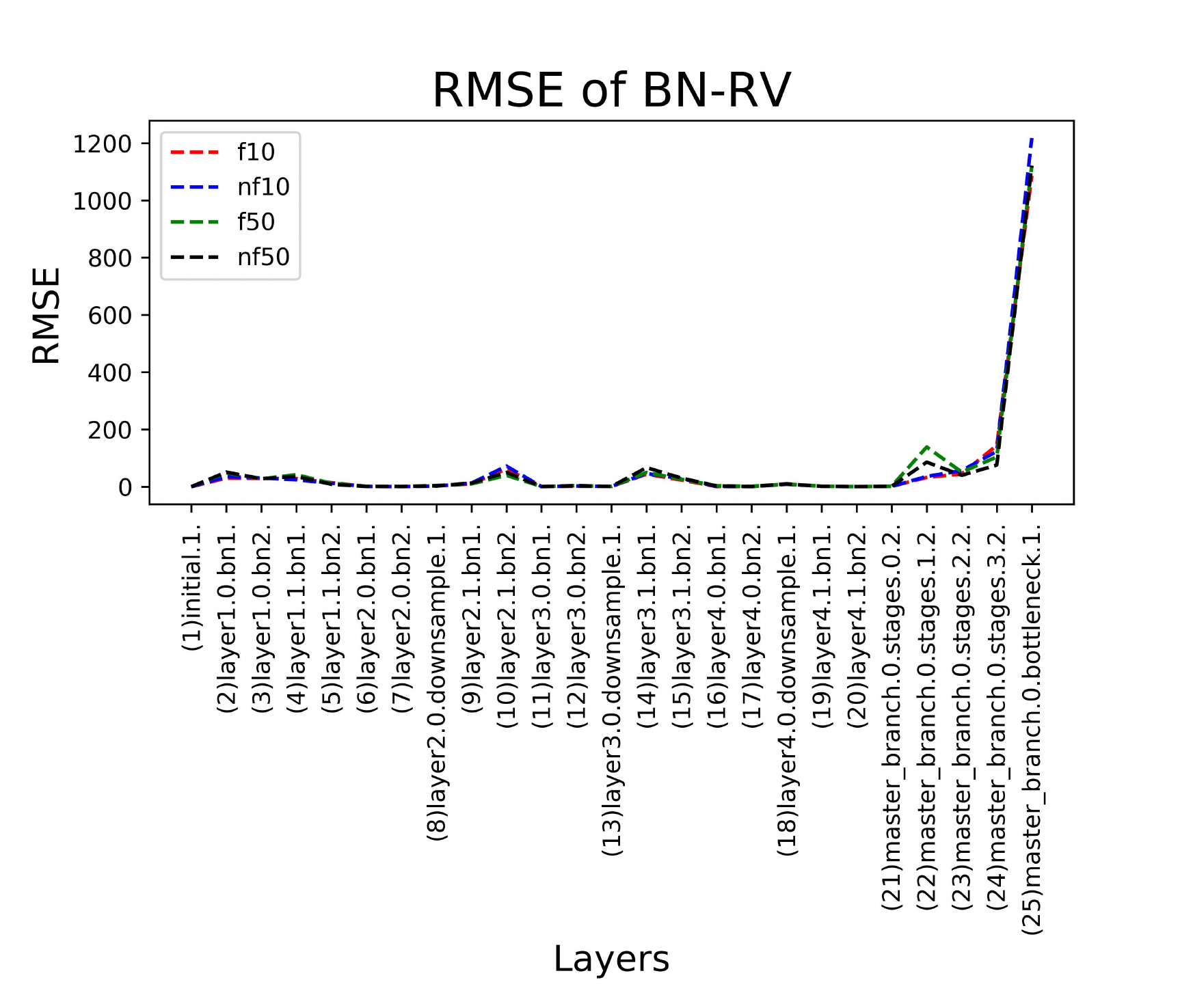}

\includegraphics[scale=0.138]{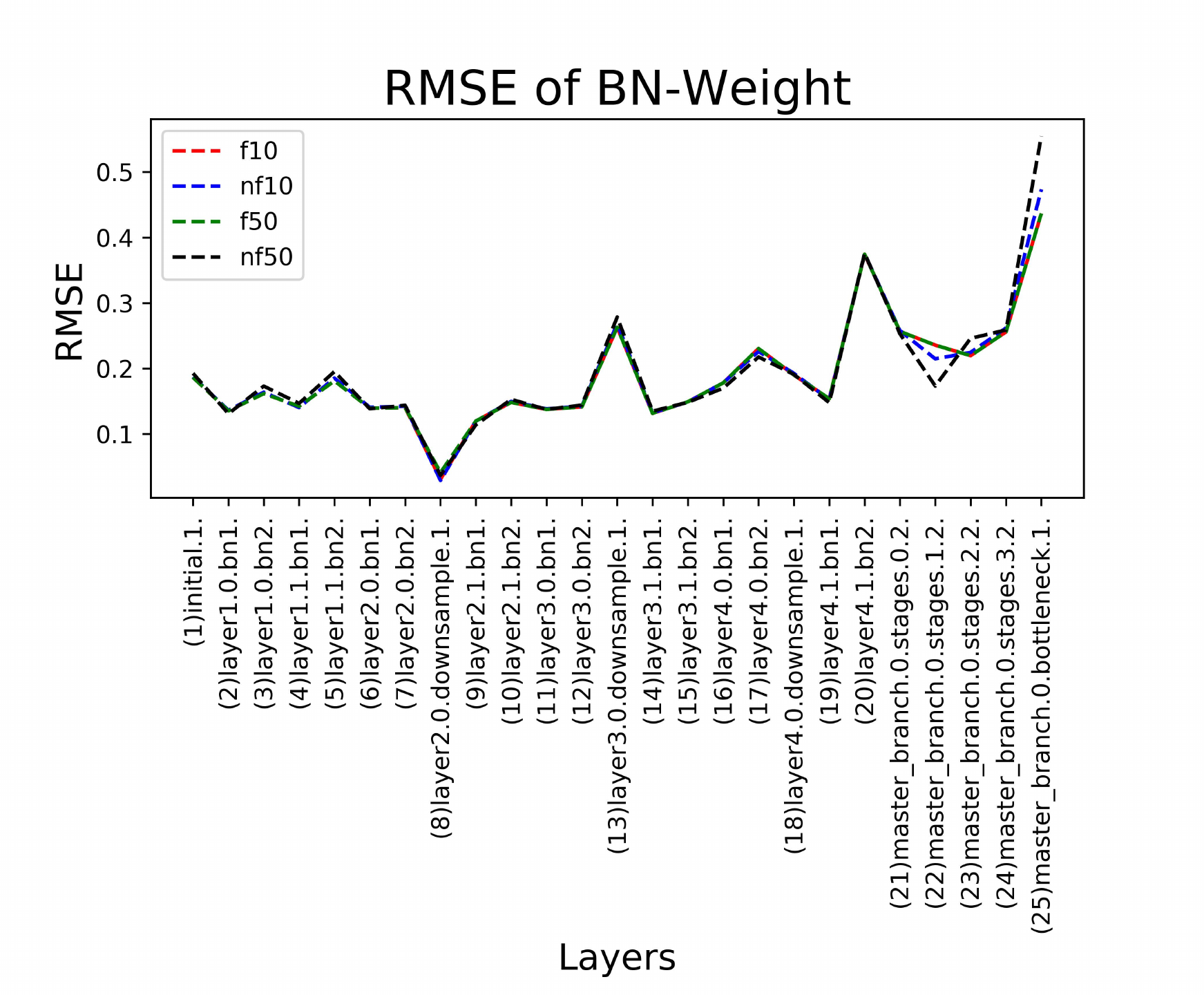}
\includegraphics[scale=0.14]{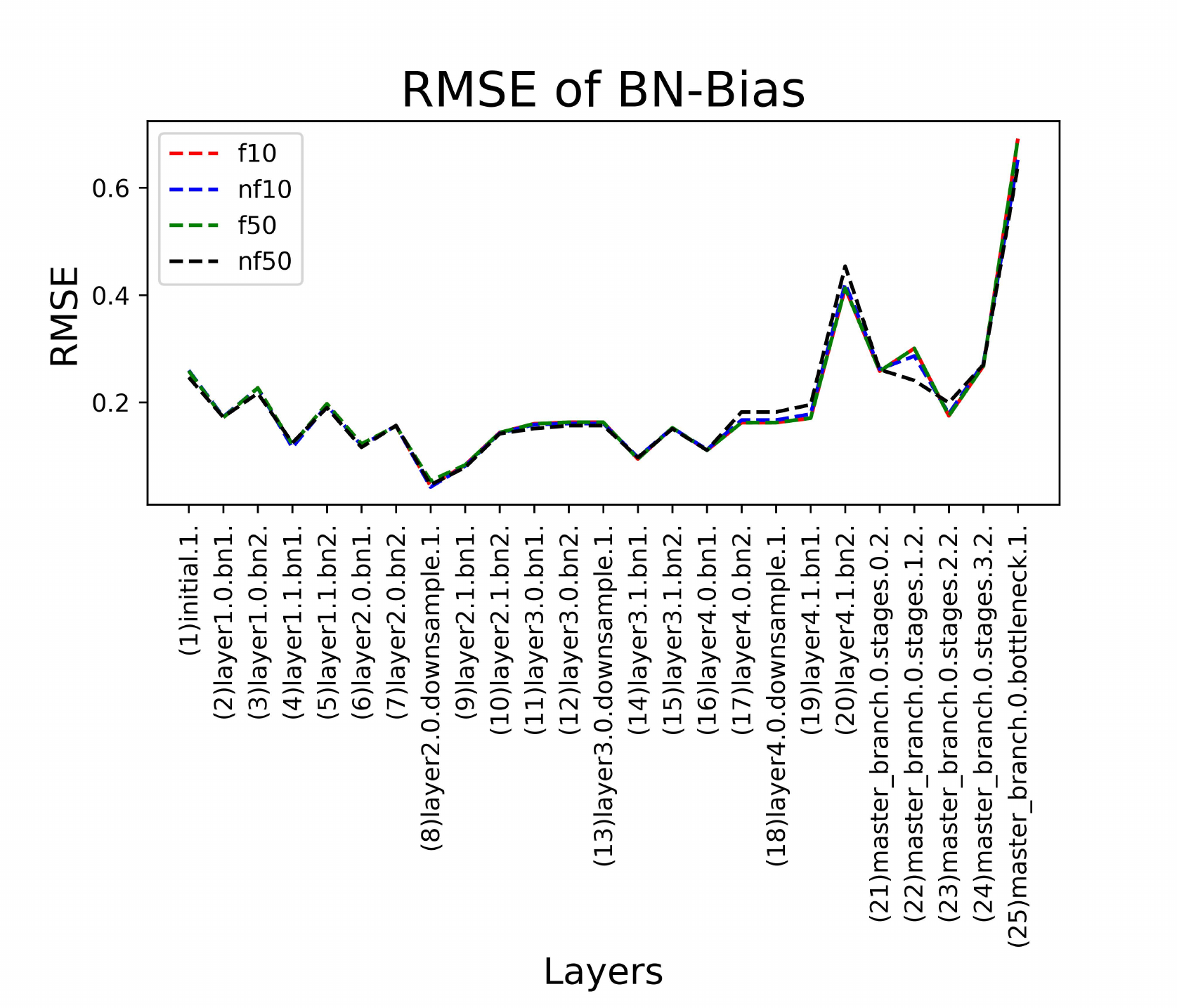}

\includegraphics[scale=0.15]{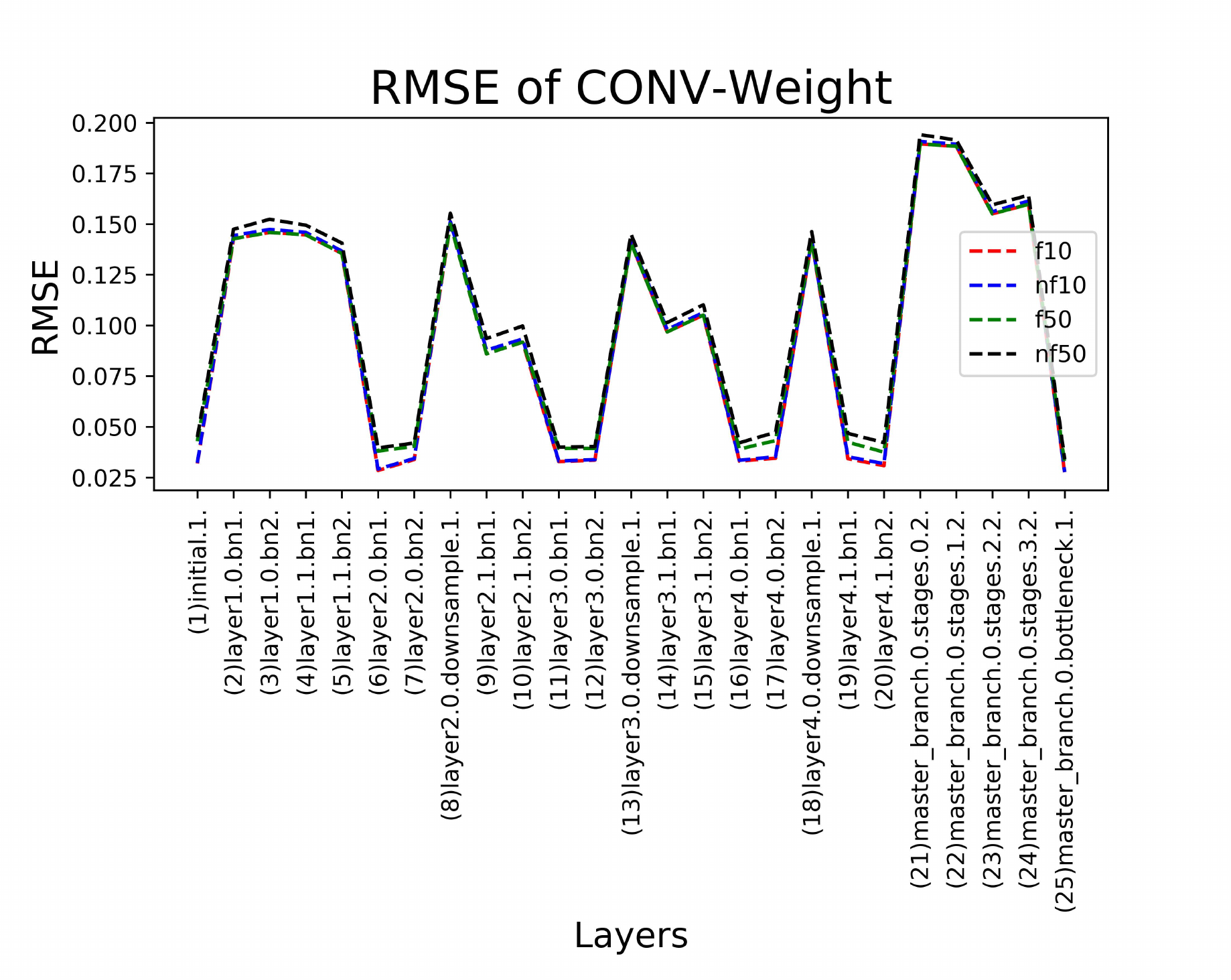}
\includegraphics[scale=0.15]{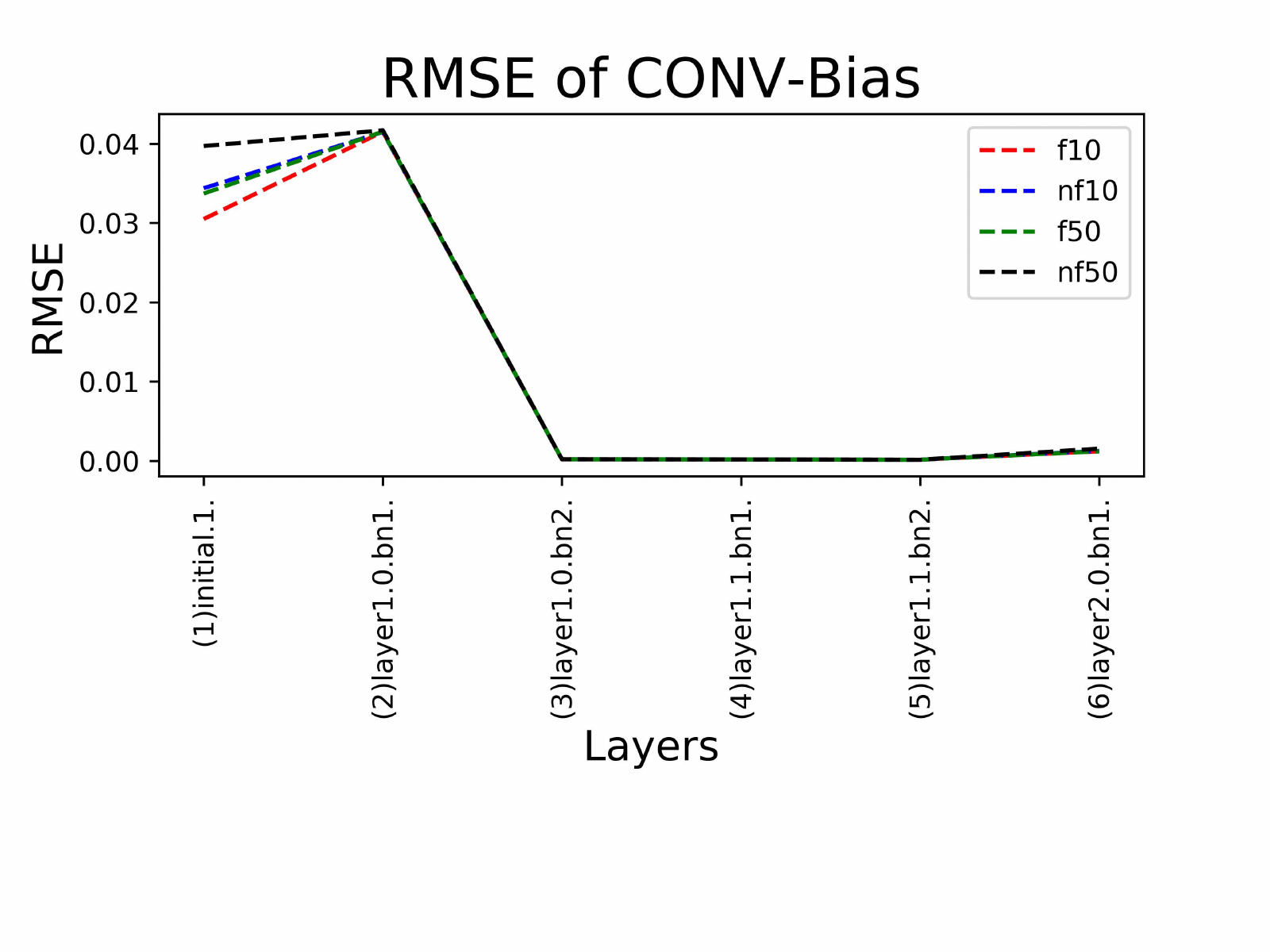}

\caption{PSPNet: After RVSC image auto-encoder model parameters are reused by ACDC image segmentation models, the difference of parameters of corresponding layer of each model. The horizontal axis represents the layers of the model, the vertical axis represents the RMSE values, and the title represents the variable whose RMSE values are computed. The icons represent different models. n: Do not load pretrained parameters; f: Freeze loaded pre-trained parameters; nf: Do not freeze loaded pre-trained parameters.10,50: The number of patients is used for training.
}

\label{pspnet-RMSE_T1_ACDC_RVSC}
\end{figure} 

Fig.16 shows that the corresponding layer parameters of the model loaded with pre-training parameters have a small difference compared with those without. If T1 nf50 (the model reuses the T1 image auto-encoder parameter and uses 50 samples of ACDC as the training set) is regarded as the ideal model for ACDC image segmentation. Fig17 shows that there is little difference between the model loaded with RVSC pre-training parameters and the corresponding layers(ResNet module does not load the pre-training parameters) of the ideal model .This paper proposes a question: for a single solution problem (i.e. the solution is unique for the segmentation task), is the solution unique for the same CNN model?(That is, if there are multiple training models and the results are very close, then whether the difference of parameters at each layer of each model must be very small).Based on the current experimental results, this paper concludes that when the model is fixed and the data set is fixed, the solution of the model with the best result is unique for a single solution problem.

\section{Discussion}
Based on the Experiment module UNet and PSPNet parameter replacement results, it was found that:  

The Conv layer in PSPNet obviously presents a regular change.Comparing the results of UNet network, there may be two reasons for this result:  

A. The convolution layers of feature number changes or the convolution layers of the down-sampling changes the generality of adjacent layer parameters.  

B. ResNet has the ability to generalize features, rather than assume the above.  

In order to explore the reasons, relevant experiments were carried out in the supplementary materials, and the conclusion was that ResNet had the ability to generalize the features. This generalization is believed to be derived from the special structure of ResNet, in which UNet has shot-connection structure. PSPNet does not, so in the process of feature learning, it is necessary to try to preserve the complete features of the original image, and such preservation of the original features corresponds to the identity mapping of ResNet. In the learning process of network, the features of original images should be preserved, and the preservation of such features is the general source of network.

\section{Conclusion}

To sum up, this paper proposes an approach to explore the reusability of network parameters, which is verified theoretically and experimentally. Network parameters are reusable for two reasons:  

1. Network features are general.  

2. There is little difference between the pre-training parameters and the ideal network parameters. If the difference is large, it is not conducive to parameter reuse.  

According to the experiment in this paper, the following conclusions can be drawn:  

1. RM of BN layers have a great influence on the final result in the first layer.  

2. RM and RB change the data distribution by changing the data shift, thus affecting the network results. RV and RW affect the network results by changing data scaling and data distribution.RM and RV of the BN layer are more important to the final result than RW and RB, which are almost entirely reusable.  

3. The model training parameters are correlated with different tasks in the same data set and network.  

4. With the same network and the same type of data set, model training parameters have different correlations according to different tasks, which requires specific parameter analysis.  

5. Transfer learning may not necessarily improve the results. If there is a big difference between the pretrained parameters and the ideal model parameters of this dataset (or the parameters that make the model results better), the results may be reduced.  

6. ResNet module has the ability to remove specialization of convolution layers (task specialization and data set specialization), so ResNet is more suitable for parameter reuse tasks.

\bibliography{refs}
\bibliographystyle{plain}

\end{document}